\documentclass[letterpaper]{article} % DO NOT CHANGE THIS
\usepackage{aaai23}  % DO NOT CHANGE THIS
\usepackage{times}  % DO NOT CHANGE THIS
\usepackage{helvet}  % DO NOT CHANGE THIS
\usepackage{courier}  % DO NOT CHANGE THIS
\usepackage[hyphens]{url}  % DO NOT CHANGE THIS
\usepackage{graphicx} % DO NOT CHANGE THIS
\usepackage{natbib}  % DO NOT CHANGE THIS AND DO NOT ADD ANY OPTIONS TO IT
\usepackage{caption} % DO NOT CHANGE THIS AND DO NOT ADD ANY OPTIONS TO IT
\usepackage{algorithm}
\usepackage{algorithmic}

\usepackage{newfloat}
\usepackage{listings}
\DeclareCaptionStyle{ruled}{labelfont=normalfont,labelsep=colon,strut=off} % DO NOT CHANGE THIS
\floatstyle{ruled}
\newfloat{listing}{tb}{lst}{}
\floatname{listing}{Listing}
\newcommand{\ie}{\textit{i}.\textit{e}.}
\newcommand{\eg}{\textit{e}.\textit{g}.}

\usepackage{graphicx, subfig}
\usepackage{amsmath}
\usepackage{amsfonts}
\usepackage{dsfont}
\usepackage{enumitem}
\usepackage{multirow}
\usepackage{makecell}
\usepackage{subfloat}
\usepackage{xcolor}
\usepackage[colorlinks=true]{hyperref} % -- This package is specifically forbidden

\title{Contrastive Masked Autoencoders for Self-Supervised Video Hashing}
\author {
    Yuting Wang\textsuperscript{\rm 1,3\equalcontrib},
    Jinpeng Wang\textsuperscript{\rm 1,3\equalcontrib},
    Bin Chen\textsuperscript{\rm 2}\thanks{Corresponding author.},
    Ziyun Zeng\textsuperscript{\rm 1,3},
    Shutao Xia\textsuperscript{\rm 1,3}
}
\affiliations {
    \textsuperscript{\rm 1} Tsinghua Shenzhen International Graduate School, Tsinghua University\\
    \textsuperscript{\rm 2} Harbin Institute of Technology, Shenzhen\\
    \textsuperscript{\rm 3} Research Center of Artificial Intelligence, Peng Cheng Laboratory\\
    \{wangyt22, wjp20, zengzy21\}@mails.tsinghua.edu.cn, chenbin2021@hit.edu.cn, xiast@sz.tsinghua.edu.cn
}

\usepackage{bibentry}
\begin{document}

\maketitle

\begin{abstract}
    Self-Supervised Video Hashing (SSVH) models learn to generate short binary representations for videos without ground-truth supervision, 
    facilitating large-scale video retrieval efficiency and attracting increasing research attention. The success of SSVH lies in the understanding of video content and the ability to capture the semantic relation among unlabeled videos. Typically, state-of-the-art SSVH methods consider these two points in a two-stage training pipeline, where they firstly train an auxiliary network by instance-wise mask-and-predict tasks and secondly train a hashing model to preserve the pseudo-neighborhood structure transferred from the auxiliary network. This consecutive training strategy is inflexible and also unnecessary. In this paper, we propose a simple yet effective one-stage SSVH method called ConMH, which incorporates video semantic information and video similarity relationship understanding in a single stage. To capture video semantic information, we adopt an encoder-decoder structure to reconstruct the video from its temporal-masked frames. Particularly, we find that a higher masking ratio helps video understanding. 
    Besides, we fully exploit the similarity relationship between videos by maximizing agreement between two augmented views of a video, 
    which contributes to more discriminative and robust hash codes. 
    Extensive experiments on three large-scale video datasets (\ie, FCVID, ActivityNet and YFCC) indicate that ConMH achieves state-of-the-art results. Code is available at \url{https://github.com/huangmozhi9527/ConMH}.
\end{abstract}

\section{Introduction}
    \begin{figure}
      \centering
      \includegraphics[width=\linewidth]{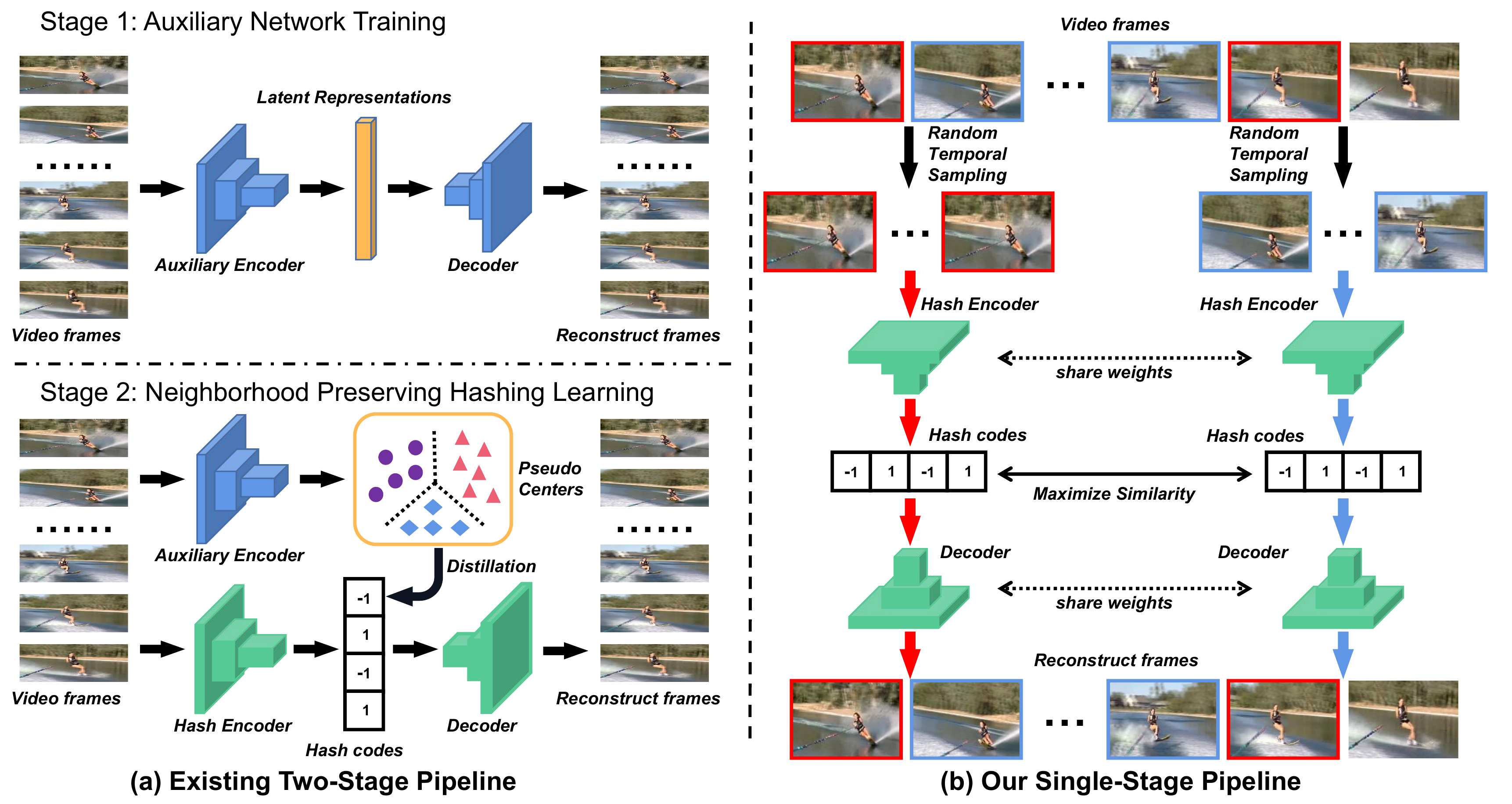}
      \caption{Illustration of the current two-stage SSVH training pipeline and our single-stage pipeline. Previous methods (a) firstly train an auxiliary network to provide the pseudo neighborhood structure, which is then utilized to supervise the relationship learning within videos. Instead of neighborhood preserving, our method (b) adopts contrastive learning to explore the similarity structure between videos.}
      \label{fig1}
    \end{figure}
    
    With the development of society, videos from various social media and search engines are increasingly abundant. Therefore, to pursue a better balance between retrieval performance and efficiency, hashing has been widely applied to large-scale video retrieval. The hashing methods map high-dimensional data to compact binary codes \cite{gao2015scalable, luo2018fast, cui2021two, SwinFGHash, HuggingHash}, making the retrieval process high-speed and require low memory footprint. Besides, due to the large amount of video data, manual annotation is time-consuming and labor-intensive. Therefore, Self-Supervised Video Hashing (SSVH) has attracted increasing attention to training the video hashing model in an unsupervised manner.

    Video retrieval needs to fully explore the temporal dynamics in video itself and exploit the similarity relationship between videos. In the past few years, there has been a lot of research \cite{zhang2016play, li2019neighborhood, li2021self, hao2022unsupervised, MAGRH} using sequential (\eg, RNN, LSTM, and Transformer, etc.) or graph-based models to solve self-supervised video hashing problem. Some of these methods only utilize a reconstruction task to let the model learn video semantic information and ignore the similarity relationship between videos, such as SSTH \cite{zhang2016play}, JTAE \cite{li2017jointly} and UDVH \cite{wu2017unsupervised}, leading to poor retrieval performance. Other methods, \eg, SSVH \cite{song2018self}, NPH \cite{li2019neighborhood}, BTH \cite{li2021self}, MCMSH \cite{hao2022unsupervised} and MAGRH~\cite{MAGRH}, learn to preserve the semantic neighborhood structure among videos into binary representations. However, these methods need to generate a dataset-covering pseudo neighborhood structure to supervise the relationship learning among videos, as shown in Figure \ref{fig1} (a). Such a two-stage framework makes the training process impractical. In addition, most SSVH methods take the complete video input, which contains much redundant information between frames. Such redundant information will be utilized when the model reconstructs the video frames. For instance, when the model has reconstructed the frame at a certain time, the next frame can be easily obtained by adding dynamic change to the current frame. In this case, the model can solve the reconstruction task easily without understanding the semantic video information well. 

    In this paper, we propose a simple yet effective one-stage framework called ConMH for self-supervised video hashing (SSVH), as shown in Figure \ref{fig1} (b). Similar to past methods, we adopt an encoder-decoder structure for reconstruction. However, to make the model better understand the semantic information of the video, we only operate the encoder on highly temporal-masked video frames to generate hash codes during training. By doing so, the input becomes sparse. With such sparse input, our model has to fully exploit the long-distance dynamics among video frames to solve the reconstruction task, therefore generating informative hash codes. And the temporal-masking operation is a way of data augmentation, making our model more robust. Besides, with a high masking ratio, the encoder only takes a small subset of the original full frames, accelerating the training phase. To simultaneously consider the similarity relationship between videos, we combine an instance-discriminative task with the above reconstruction task in a single stage. Inspired by contrastive learning in self-supervised representation learning, we think two different views of the same video should have the same hash codes. Specifically, we consider random temporal masking as a special data augmentation designed for contrastive learning and sample two small non-overlapping subsets from the full video frames. These two subsets can be seen as two different augmented views of the same video. Then we feed these two subsets to the encoder to generate two hash codes. Finally, a contrastive learning method is utilized to maximize the similarity between these two hash codes. We choose debiased contrastive learning as our contrastive learning method, due to its ability to mitigate the adverse effect of false negative samples. 

    We conducted extensive experiments on three large-scale video datasets FCVID \cite{jiang2017exploiting}, ActivityNet \cite{caba2015activitynet}, YFCC \cite{thomee2015new}. The experimental results prove that ConMH achieves state-of-the-art results under all metrics. In particular, on FCVID dataset with 16 bits hash codes, the mAP@5 and mAP@100 of ConMH are ahead of the past SOTA MCMSH \cite{hao2022unsupervised} by 6.1\% and 30.0\% respectively. 

    Overall, the contributions of this paper are as follows:
    \begin{itemize}[leftmargin=*]
    \item Beyond two-stage prior arts, we contribute ConMH, a simple yet effective one-stage learning paradigm for self-supervised video hashing. 
    \item We present a natural and general idea in ConMH: the temporally masked autoencoder enhances intrinsic understanding within each video, while debiased contrastive learning helps to perceive discriminative semantic relation among videos; these two objectives are complementary in a unified hashing learning framework.
    \item Experimental results on three datasets (\ie, FCVID, ActivityNet, YFCC) demonstrate the superiority of ConMH.
    \end{itemize}

\section{Related Works}

    \textbf{Self-Supervised Video Hashing.} Hashing retrieval methods map data into binary hash codes, which can perform efficient and low storage requirement retrieval because of the fast bit $XOR$ operations in the Hamming space. In the early research, most methods directly regarded a video as a set of disordered images. They adopted hashing technologies that already exist in image hashing, such as iterative quantization \cite{gong2012iterative}, spectral hashing \cite{weiss2008spectral} and multiple feature hashing (MPH) \cite{song2011multiple}. However, performance of these methods is not good because they ignore the temporal dynamics in the video. In recent years, with the development of deep neural networks, many hashing methods for video retrieval have been proposed. For example, \citet{zhang2016play} proposed SSTH, an end-to-end framework to capture the correlations between video frames. \citet{li2017jointly} thought both the static visual appearance and temporal patterns of videos contain important discriminative information and concentrated on combining them to generate more effective hash codes. \citet{wu2017unsupervised} proposed to apply a balanced rotation to video-specific features widely distributed in low-dimensional space, which can balance the variance of dimensions. \citet{song2018self} found that the reconstruction objective alone does not perform the retrieval task well, and attention needs to be paid to instance similarity. NPH \cite{li2019neighborhood} proposed a neighborhood attention mechanism to enable the encoder to selectively encode frames in the video that contain rich spatial-temporal neighborhood information. \citet{li2021self} improved NPH's method of generating similarity matrices and applied the transformer architecture to self-supervised video hashing for the first time. And \citet{hao2022unsupervised} enhances the discrimination of hash codes by utilizing self-gating modules. However, most neighborhood-preserving methods use a two-stage training strategy, which is not practical. In this paper, we design a simple yet effective one-stage masked autoencoders framework (ConMH), which only takes highly temporal-masked frames as input during training and considers the similarity structure between videos in a single stage.  

    \textbf{Masked Autoencoders.} In the past few years, contrastive learning \cite{chen2020simple, he2020momentum, chen2021exploring, grill2020bootstrap, caron2020unsupervised, dong2021peco} has become mainstream in the field of self-supervised learning. It models instance similarity and dissimilarity between two or more views of the same data example. Recently, there have been a lot of approaches \cite{bao2021beit, he2021masked, chen2022context, xie2021simmim, assran2022masked} that utilize masking operations for self-supervised learning. BEiT \cite{bao2021beit} proposed a novel training strategy for image pretraining called masked image modeling (MIM), which is a reconstruction task. \citet{he2021masked} pointed out that contrastive and related methods strongly depend on data augmentation, and proposed a novel masked autoencoders framework called MAE, providing a new idea for visual self-supervised learning. Besides, MAE found that a high masking ratio benefits representation learning. However, MAE ignores the similarity structure within data while only applying the reconstruction task, which is disadvantageous for some downstream tasks, such as classification and retrieval. Our ConMH combines an instance-discriminative task with the reconstruction task to simultaneously consider the visual semantic information of video and the similarity structure between videos.

\section{The Proposed ConMH Framework}

    \begin{figure*}[t]
      \centering
      \includegraphics[width=\linewidth]{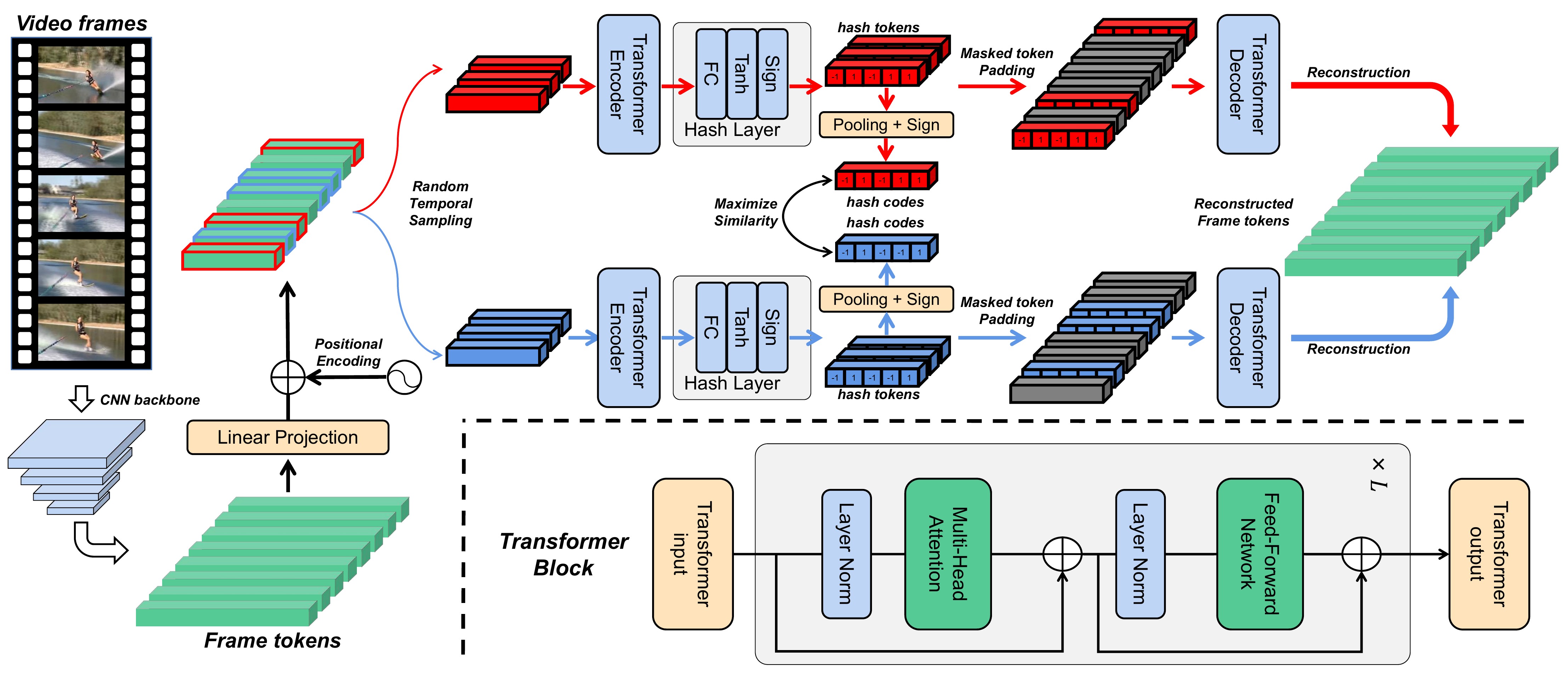}
      \caption{\textbf{Our ConMH architecture.} Given the uniformly downsampled frames of a video, we use an ImageNet-pretrained CNN Backbone to extract frame tokens. Then we randomly sample two small frame token subsets (marked in red/blue), which will be fed to a transformer encoder and a hash layer to generate two hash codes. A debiased contrastive loss is adopted to maximize the similarity between two hash codes, as they are sampled from the same video. Finally, we use a transformer decoder to reconstruct the original video frame tokens separately from hash tokens (red/blue square) and masked tokens (grey square).}
      \label{fig2}
    \end{figure*}

    In this section, we present our proposed ConMH framework. Given a set of $N$ videos, ConMH aims to learn a nonlinear hash function that maps each video to a compact binary code. Specifically, a video is represented by CNN features with $M$ frames $ \{ v_i^m \}_{m=1}^{M} \in \mathbb{R}^{M \times  d} $, where $d$ is the frame feature dimension  and $i$ is the video index. Then we feed $ \{ v_i^m \}_{m=1}^{M} $ to the ConMH encoder to get its binary code $b_i \in \{ -1, 1 \}^k$, where $k$ is the code length.

    \subsection{Video Semantic Information Understanding}
        ConMH is a simple yet effective masked autoencoders framework for self-supervised video hashing. To better understand the semantic information of the video, ConMH reconstructs the original video given its highly temporal-masked frames. Following MAE \cite{he2021masked}, the ConMH architecture is asymmetric. Specifically, the encoder of ConMH only takes a small randomly sampled subset of the original frames as input to extract latent representations and binary hash tokens. And the decoder of ConMH reconstructs the original video frames from generated hash tokens and masked tokens, illustrated in Figure \ref{fig2}.

        \textbf{Masking.} To make the model better understand the semantic information of the video, we adopt a highly temporal-masking strategy. Specifically, given the frames $ \{ v_i^m \}_{m=1}^{M} $ of a video, we randomly sample a small subset of the frames for training and mask the remaining ones. Typically, a higher masking ratio means less available information the model can get, leading to a more challenging reconstruction task. Given these intuitions, the model can learn high-quality representations and generate effective hash codes by a reasonable self-supervised pretext task. Besides, with a high masking ratio, the encoder only operates on a small subset of an input, accelerating the training phase.

        \textbf{ConMH Encoder.} Our encoder is composed of a standard Transformer \cite{vaswani2017attention} and a hash layer. The multi-head attention module in Transformer is able to capture the temporal information and correlations among video frames, which is critical for video understanding. Different from the setting in \cite{vaswani2017attention}, we only use a small subset (\eg, 25\%) of the full video frames as the transformer input during training. Through multiple transformer blocks, we can obtain the latent representations $ \{ h_i^m \}_{m=1}^{\tilde{M}}$ for each video, where $\tilde{M}$ is the frame number of the sampled subset. To obtain compact binary code, we further adopt a hash layer to generate binary hash tokens $\{ t_i^m \}_{m=1}^{\tilde{M}}$. Finally, we generate a binary code for each video by operating the mean pooling and sign function on its binary hash tokens. Since the sign function is not differentiable, we employ the approach in \cite{li2021self} to realize the training phase in an end-to-end manner.

        \textbf{ConMH Decoder.} Our decoder is composed of another series of Transformer blocks. Different from the encoder, our decoder takes binary hash tokens and corresponding masked tokens as input. Positional encoding is also inserted to identify the masked token with their corresponding frames. The decoder is only used during the training phase, so we can flexibly design the decoder architecture. Following the asymmetric nature of MAE \cite{he2021masked}, we design a small Transformer decoder, with which the encoder has to learn high-quality feature representations and hash codes to complete the reconstruction task.

        \textbf{Reconstruction Task.} After ConMH encoder and decoder module, we can obtain a full set of reconstruction frames $ \{ \tilde{v}_i^m \}_{m=1}^{M} $. Following MAE, we only reconstruct the masked frames using corresponding reconstruction frames $ \{ \tilde{v}_i^m \}_{m=1}^{M-\tilde{M}} $. We adopt Mean Square Error (MSE) to measure the reconstruction quality as following:
        \begin{eqnarray}
          \mathcal{L}_{recon} = \frac{1}{dN(M-\tilde{M})}\sum_{i=1}^{N}\sum_{m=1}^{M-\tilde{M}} \| v_i^m - \tilde{v}_i^m \|_{2}^{2}.
        \end{eqnarray}

    \subsection{Video Similarity Relationship Understanding}

        Although the reconstruction task above can help the model to extract the discriminative visual semantic information among videos, it neglects the similarity structure between videos, which is critical for the retrieval task. In this subsection, we will discuss how to combine instance-discriminative contrastive learning with the proposed masked autoencoder in a single stage.

        \textbf{Debiased Contrastive Loss.} In ConMH, we consider random temporal masking as a data augmentation for contrastive learning and sample two non-overlapping small subsets from a single video, as shown in Figure \ref{fig1}. Then we feed these two subsets to the ConMH encoder to obtain two hash codes from different views of the same video, denoted as $b_i$ and $b_j$. For ConMH, we adopt a debiased contrastive loss  \cite{chuang2020debiased} $\ell^{de}(i,j)$ to correct the sampling bias of same-label data points:
        \begin{gather}
          \ell^{de}(i,j) = -\log \frac{e^{sim(b_i,b_j)/ \tau}}{e^{sim(b_i,b_j)/ \tau} + (2N-2)NG_{de}
          } \\
           NG_{de} =  \max \left(e^{-1/\tau},  \frac{1}{1-\rho} NG\right) \\
           NG =  \frac{1}{2N-2}\sum_{k=1}^{2N} \mathds{1}_{k\neq i,j} e^{sim(b_i,b_k)/ \tau}
          - \rho e^{sim(b_i,b_j)/ \tau}
        \end{gather}

        where $\tau$ denotes a temperature parameter and $\rho$ is the class probability indicating that two randomly sampled videos belong to the same class with probability $\rho$. Because we cannot get the actual number of categories, we treat $\rho$ as a hyperparameter. Given a mini-batch of $N$ samples, we calculate the contrastive loss over all samples below:
        \begin{gather}
          \mathcal{L}_{contra} = \frac{1}{2N} \sum_{k=1}^{N} \left(\ell^{de}(2k-1, 2k) + \ell^{de}(2k, 2k-1)\right)
        \end{gather}

        \textbf{Loss Objective.} The overall loss of ConMH for network optimization is a combination of reconstruction loss and constrastive loss as following,
        \begin{gather}
          \mathcal{L} = \mathcal{L}_{recon} + \alpha \mathcal{L}_{contra}
        \end{gather}
        where $\alpha$ is a hyperparameter that balances the above-mentioned two losses.

\section{Experiments}

    \subsection{Experimental Setup}
        \textbf{Dataset:} We evaluate our ConMH on three large-scale video datasets FCVID \cite{jiang2017exploiting}, ActivityNet \cite{caba2015activitynet} and YFCC \cite{thomee2015new}. \textbf{FCVID} contains 91,223 videos in 239 categories. Following \cite{song2018self}, we use 45,585 videos among FCVID for training and 45,600 videos for retrieval database and queries. \textbf{ActivityNet} covers 200 categories of various human activities. Due to the lack of original test labels, we use 9,722 videos as our training set and the validation set as our test set. Similar to \cite{li2021self}, we uniformly sample 1,000 videos in 200 categories as queries and the remaining 3,758 videos as retrieval database. \textbf{YFCC} contains 0.8M videos in 80 categories, which is one of the largest public video datasets available in the real world. We use 409,788 videos for training and 101,256 videos for testing. We randomly choose 1,000 videos with non-zero labels among the testing set as queries and the remaining ones as our retrieval database. 

        \textbf{Evaluation Protocols:} Following the same evaluation protocols in \cite{li2021self}, mean Average Precision at top-K retrieval results (mAP@K) is used to evaluate our ConMH performance. Videos are sorted according to their Hamming distance from the query. We evaluate the retrieval results with code lengths of 16, 32, and 64 bits.

        \textbf{Implementation Details:} Similar to BTH \cite{li2021self}, on FCVID and YFCC datasets, we uniformly sample 25 frames for each video and use VGG-16 network \cite{simonyan2014very} pre-trained on ImageNet \cite{russakovsky2015imagenet} to extract 4096-D frame features. And on ActivityNet, we uniformly sample 30 frames for each video and use ResNet50 \cite{he2016deep} pre-trained on ImageNet to extract 2048-D frame features.

        We follow the design of ViT-small \cite{dosovitskiy2020image} to build our model. Specifically, for the Transformer encoder, we set the depth as 12, multi-head number as 6, and hidden size as 256. For the Transformer decoder, we set the depth as 2, multi-head number as 3, and hidden size as 192. 

        During training, we set the batchsize as 512, masking ratio as 0.75, $\alpha$ as 1.0, $\tau$ as 0.5, $\rho$ as 0.1, and train our model for 800 epochs, 500 epochs, and 40 epochs for FCVID, ActivityNet and YFCC respectively.

        We set the initial learning rate as 0.0001, and decay it to 90\% every 20 epochs with a minimal learning rate of 0.00001. We optimize our model by using Adam optimizer algorithm \cite{kingma2014adam}. Our model is implemented in Pytorch with an Nvidia RTX2080Ti GPU.

\begin{figure}[t]
\centering
\subfloat{\includegraphics[width = 0.48\textwidth]{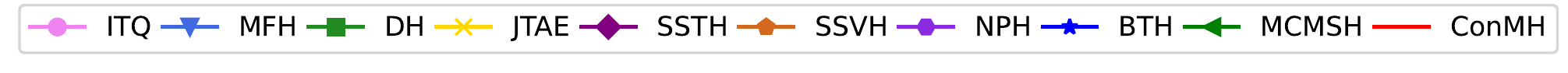}}
    \vspace{-1em}
    \setcounter{subfigure}{0}
  \subfloat[\small{FCVID 16 bits}]{\includegraphics[width = 0.1600\textwidth]{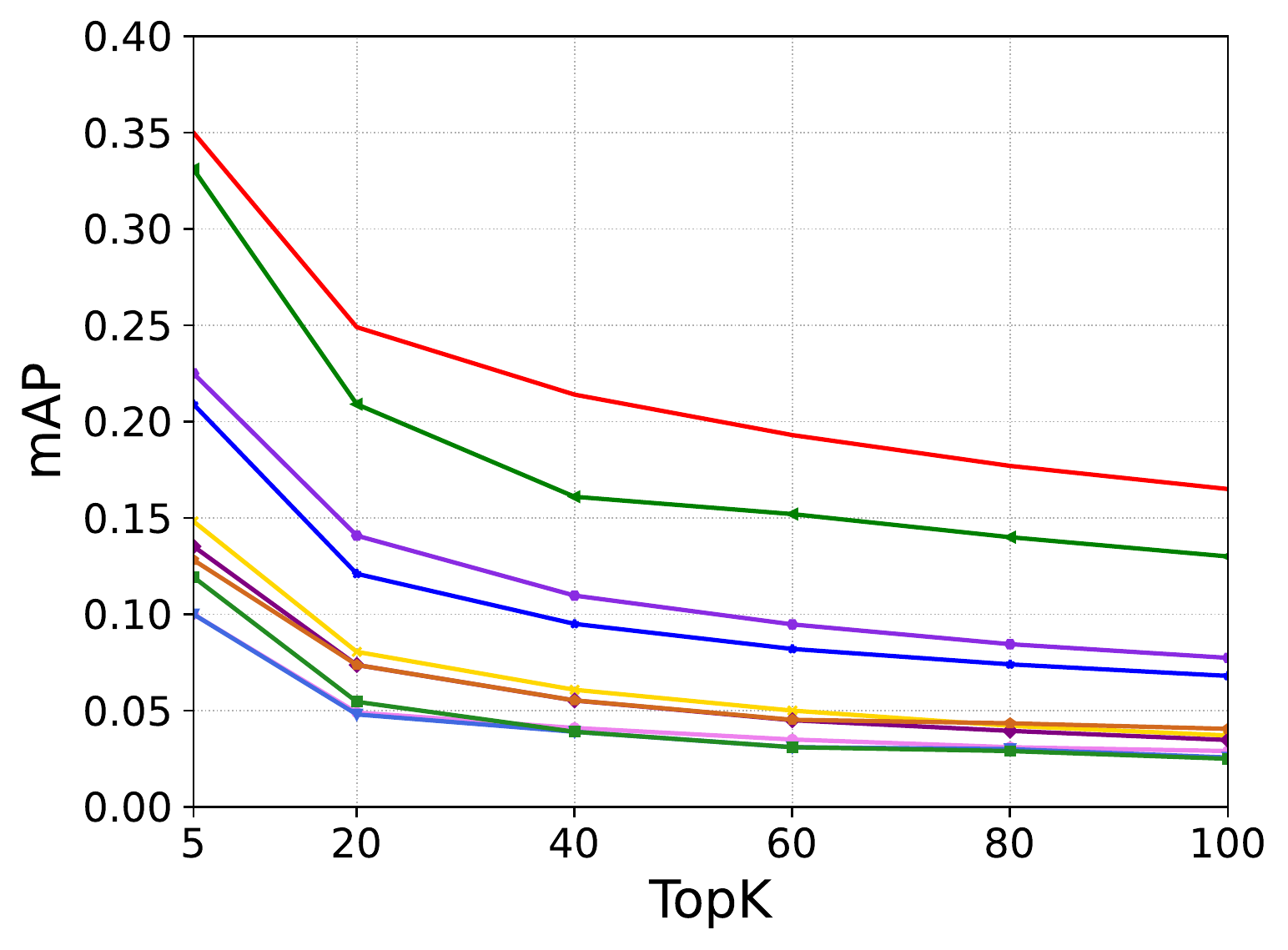}}
  \subfloat[\small{FCVID 32 bits}]{\includegraphics[width = 0.1600\textwidth]{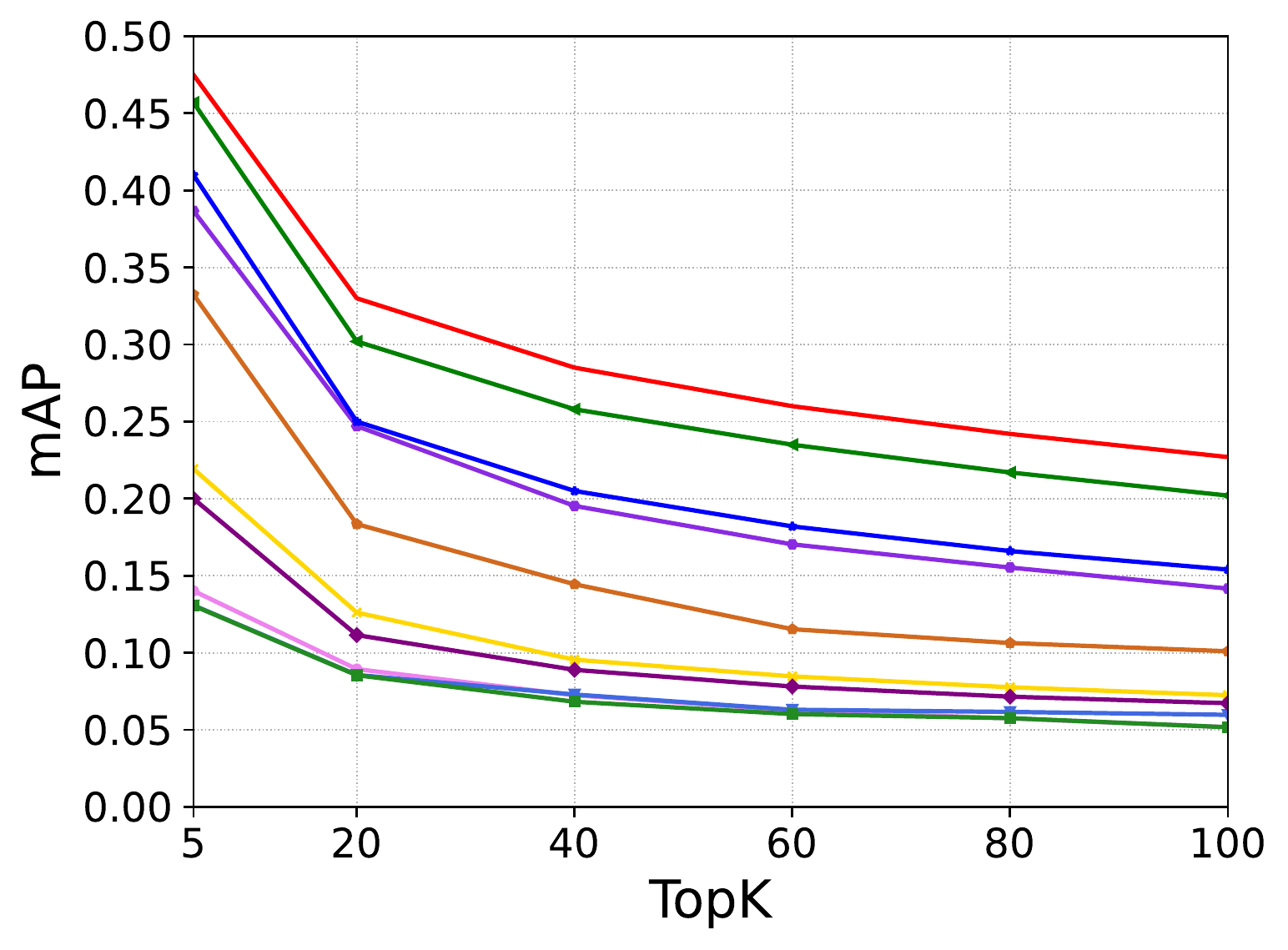}}
  \subfloat[\small{FCVID 64 bits}]{\includegraphics[width = 0.1600\textwidth]{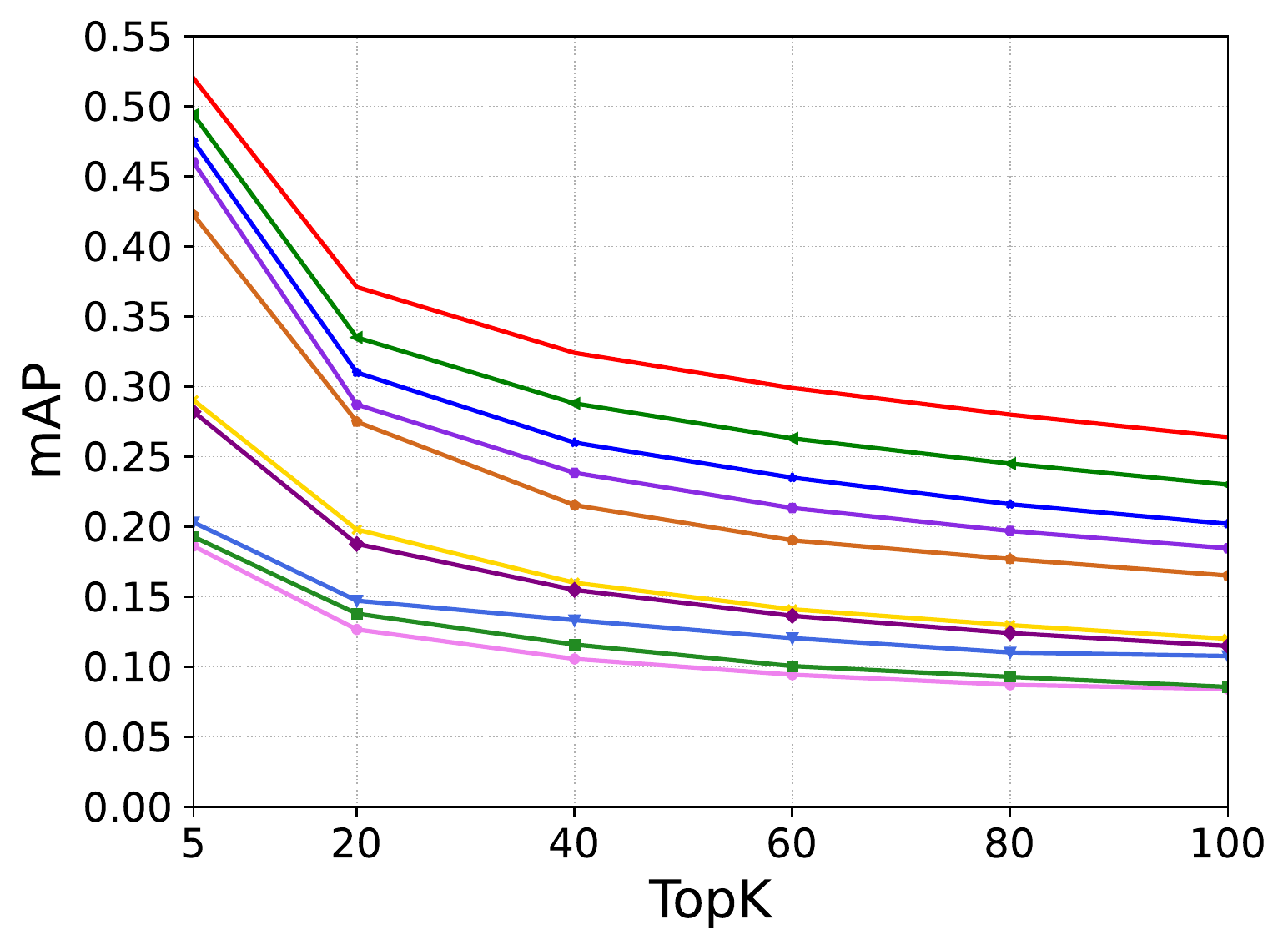}}
    \vspace{-1em}
  % \quad
  \subfloat[\small{Act-Net 16 bits}]{\includegraphics[width = 0.1600\textwidth]{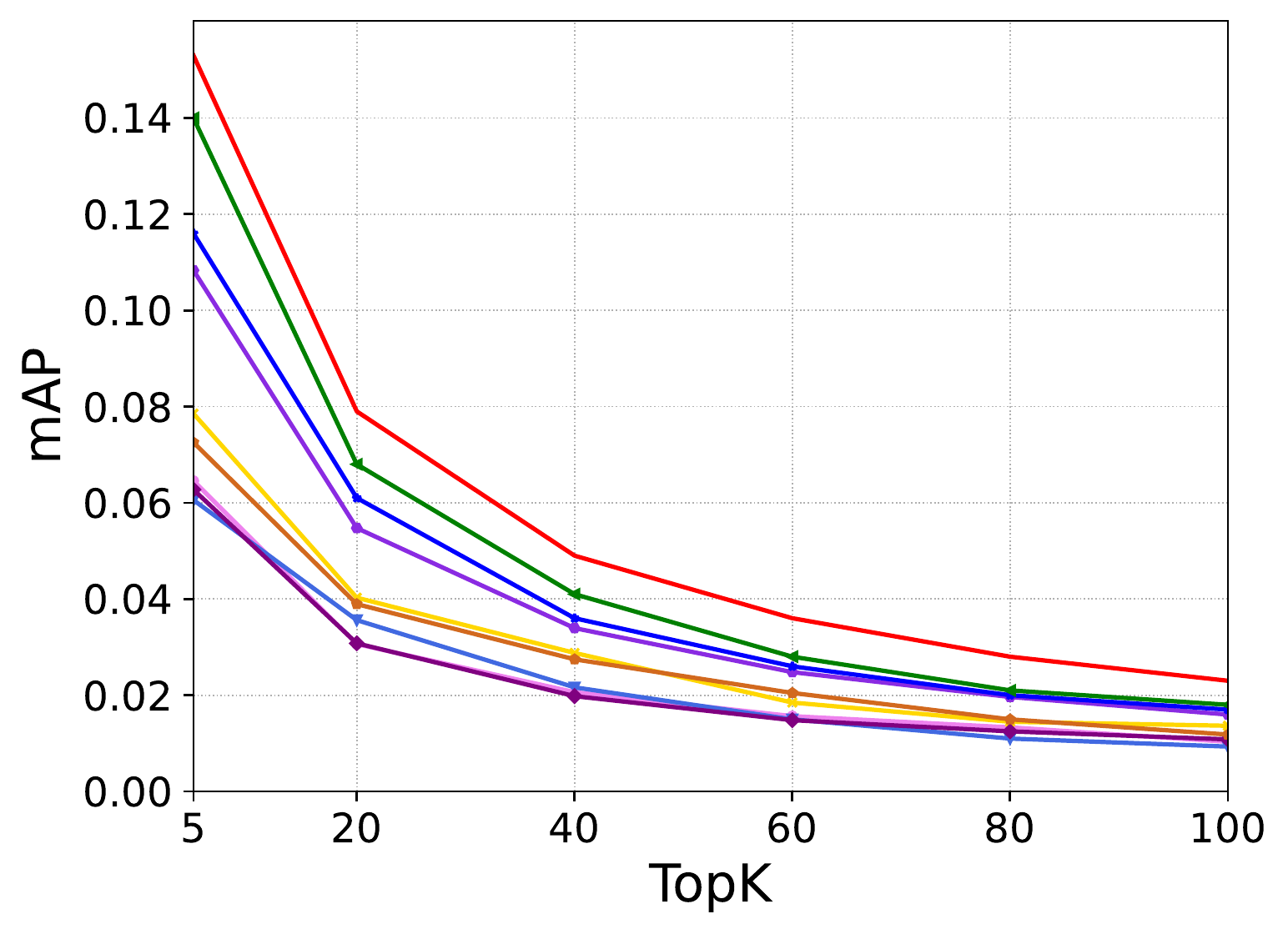}}
  \subfloat[\small{Act-Net 32 bits}]{\includegraphics[width = 0.1600\textwidth]{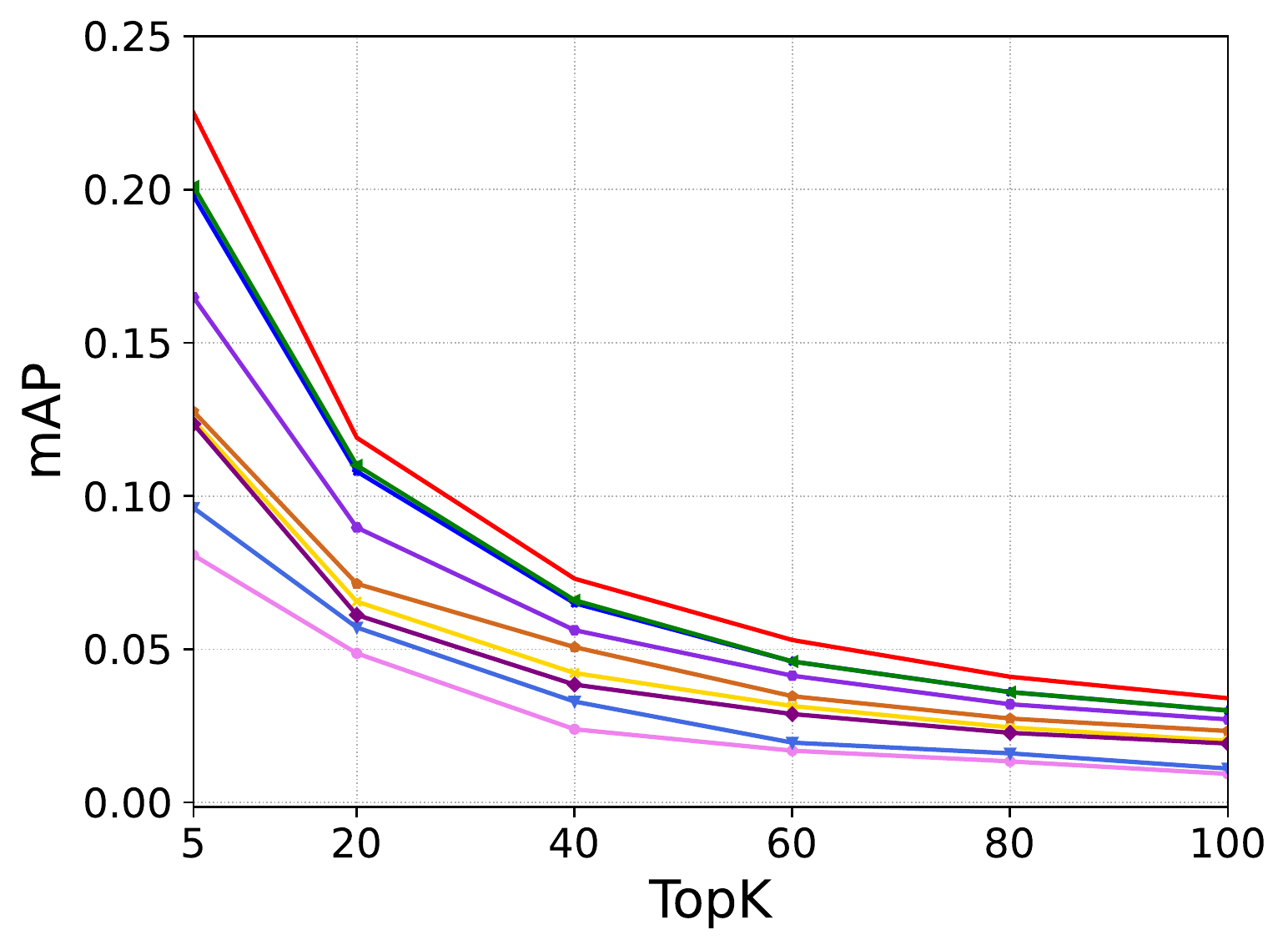}}
  \subfloat[\small{Act-Net 64 bits}]{\includegraphics[width = 0.1600\textwidth]{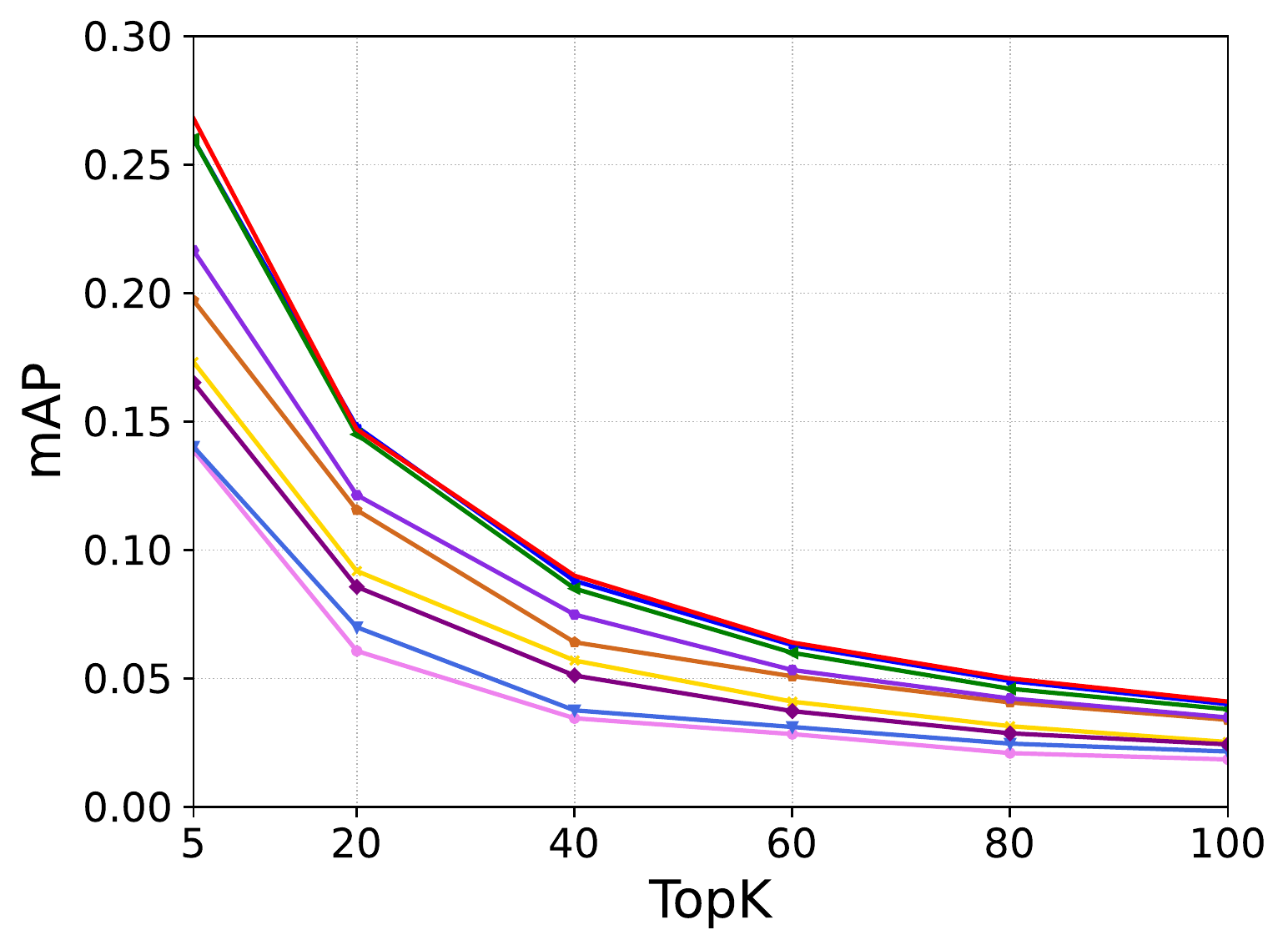}}
    \vspace{-1em}
  % \quad
  \subfloat[\small{YFCC 16 bits}]{\includegraphics[width = 0.1600\textwidth]{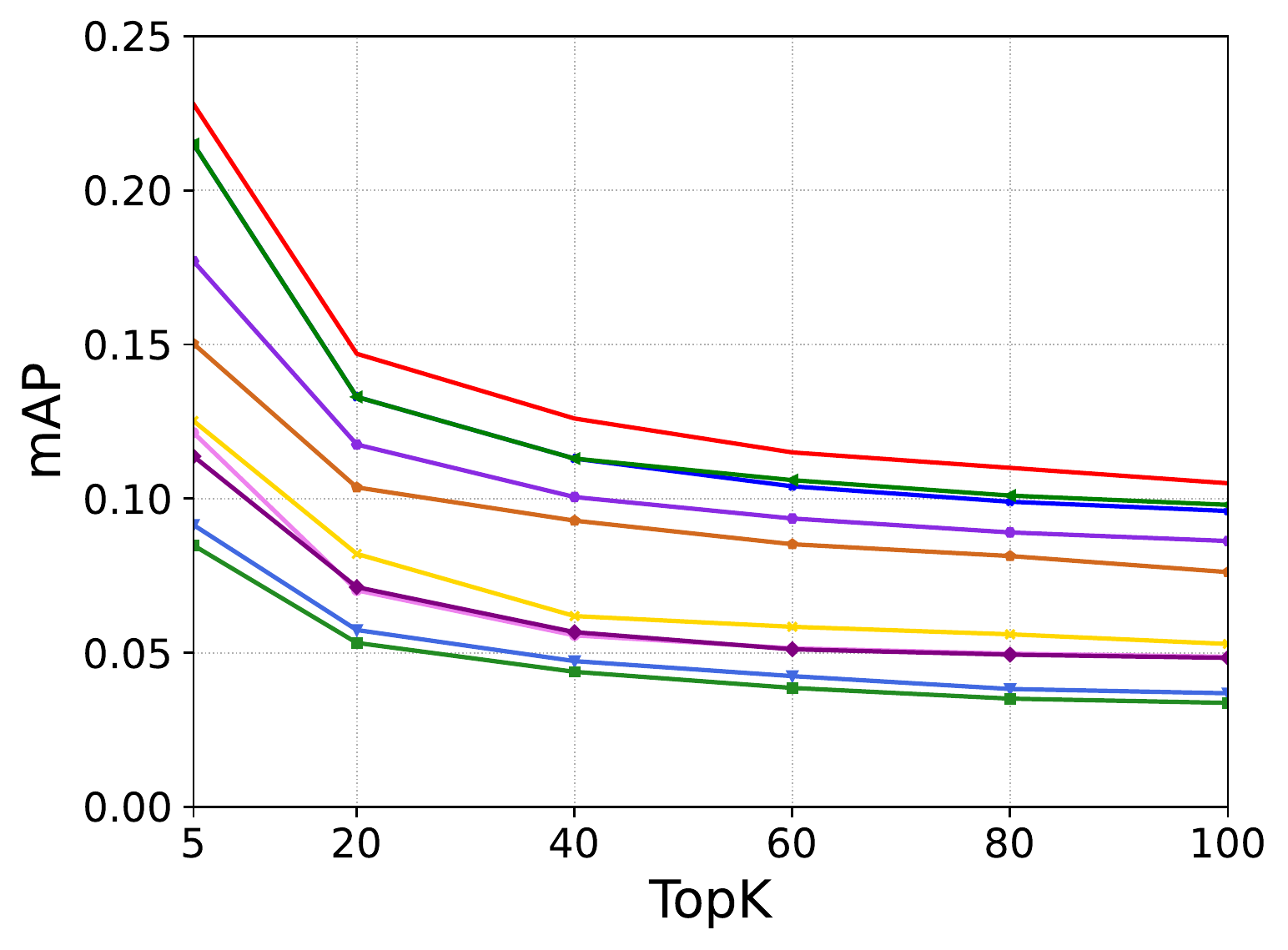}}
  \subfloat[\small{YFCC 32 bits}]{\includegraphics[width = 0.1600\textwidth]{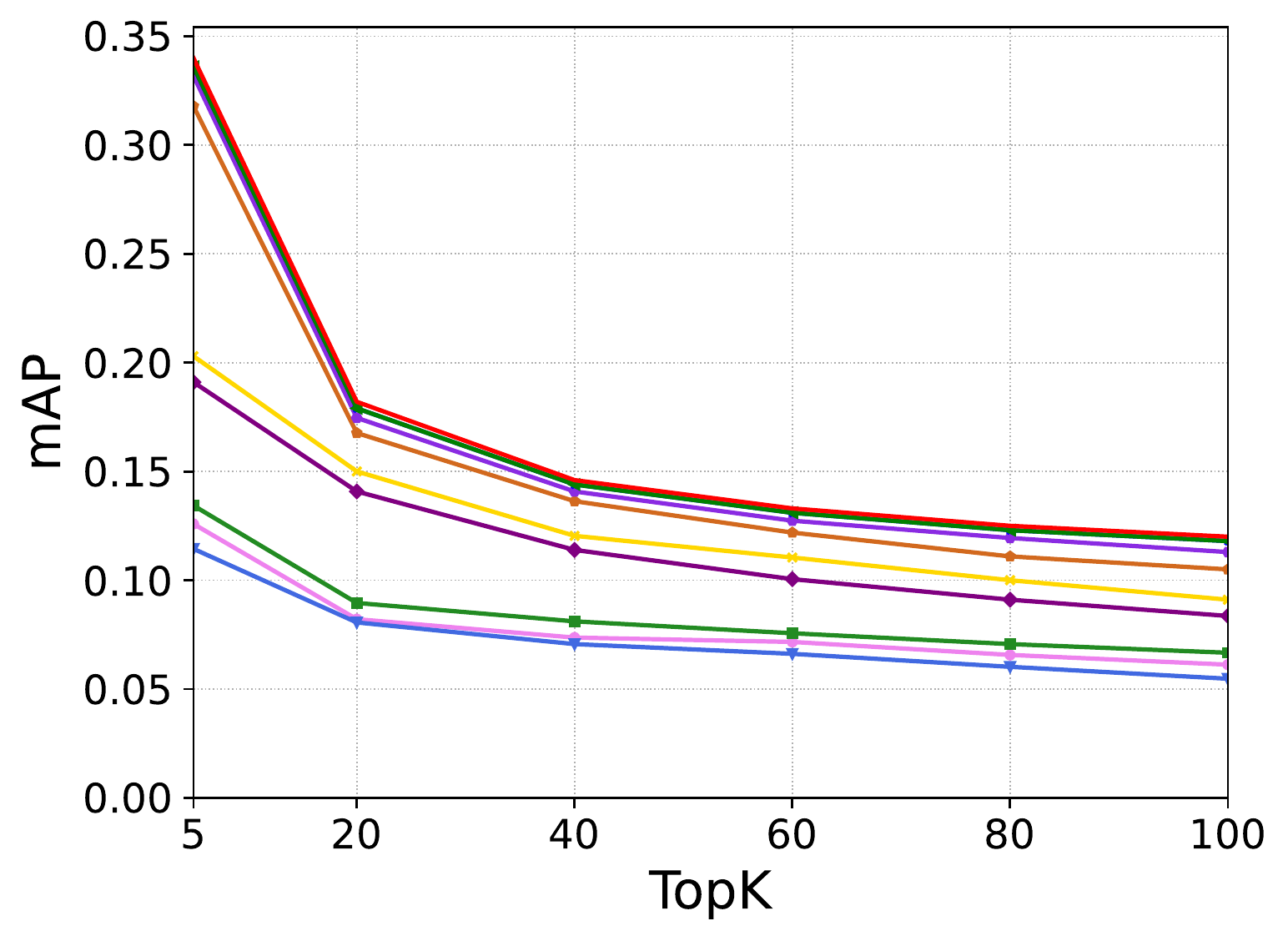}}
  \subfloat[\small{YFCC 64 bits}]{\includegraphics[width = 0.1600\textwidth]{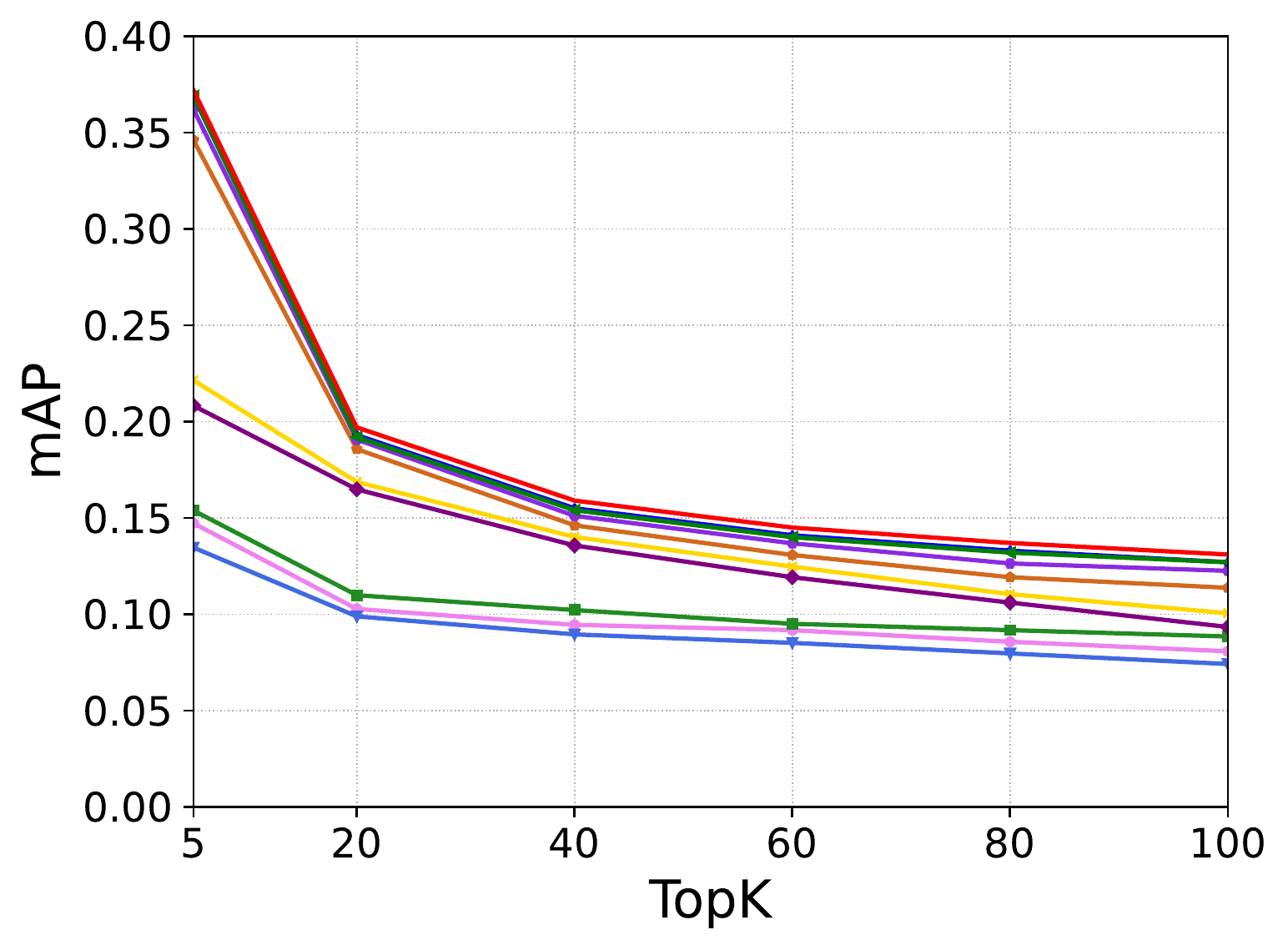}}
  \caption{Retrieval performance compared with state-of-the-arts in terms of mAP@K on FCVID (a-c), ActivityNet (d-f) and YFCC (g-i).}
  \label{sota}
\end{figure}

    \subsection{Results and Analysis}

\begin{table*}[!t]
\centering
  \begin{tabular}{c| c| c| c| c| c| c}
    % \hline
    \hline
    Method & SSTH & SSVH & NPH & BTH & MCMSH & ConMH \\
    \hline
     mAP@20 
     & 0.155 $\downarrow $ 6.3
     & 0.173 $\downarrow $ 7.8  
     & 0.180 $\downarrow $ 6.0 
     & 0.182  $\downarrow $ 5.7
     & 0.186  $\downarrow $ 3.2
     & \textbf{0.188}  $\downarrow $ 3.6  \\
     \hline
     \hline
     % \bottomrule
  \end{tabular}
  \caption{Cross-dataset mAP@20 retrieval results and corresponding performance degrades of various methods, which are trained on FCVID and tested on YFCC with 64 bits.}
  \label{crossdataset}
\end{table*}

        \textbf{Comparisons with State-of-the-arts:} We compare our ConMH with several state-of-the-art self-supervised video hashing methods: ITQ \cite{gong2012iterative}, MFH \cite{song2011multiple}, DH \cite{erin2015deep}, SSTH \cite{zhang2016play}, JTAE \cite{li2017jointly}, SSVH \cite{song2018self}, NPH \cite{li2019neighborhood}, BTH \cite{li2021self} and MCMSH \cite{hao2022unsupervised}. Following \cite{song2018self}, we extend the image hashing methods ITQ and DH to video hashing by applying them on CNN frame features. 

        As Figure \ref{sota} shows, our ConMH outperforms other methods on three large-scale video datasets with all code lengths, demonstrating its superiority in retrieval accuracy. Specifically, compared with the strongest competitor MCMSH, our ConMH surpasses it by 5.7\%, 11.4\%, and 4.7\% with 16 bits in terms of mAP@5 on FCVID, ActivityNet, and YFCC, respectively. ConMH achieves impressive retrieval results by incorporating video semantic information and video similarity relationship understanding in a single stage. These two tasks are critical for the retrieval task and can complement each other to further improve retrieval performance when combined together. 
        
        Besides, from Figure \ref{sota}, ConMH surpasses other methods by a considerable margin with low-bit hash codes, proving the superiority of ConMH in high real-time demand scenarios. In particular, as we decrease the bit length, more semantic information will be lost by the compressive hashing technique, leading to poor retrieval performance. However, instead of representation learning followed by quantization, we design hashing-oriented learning objectives to suppress the quantization loss. In this way, more semantic information will be retained by hash codes. On the contrary, when the bit-length is increasing, the quantization loss is reduced with a higher information capacity, and the hashing learning task also becomes simpler. Therefore, we can see different methods show approaching performance under 64 bits or larger bit-length settings. This phenomenon has widely occurred in previous methods, as shown in Figure \ref{sota} (f), (i).

\begin{figure}[h]
\centering
  \subfloat[\small{FCVID 16 bits}]{\includegraphics[width = 0.155\textwidth]{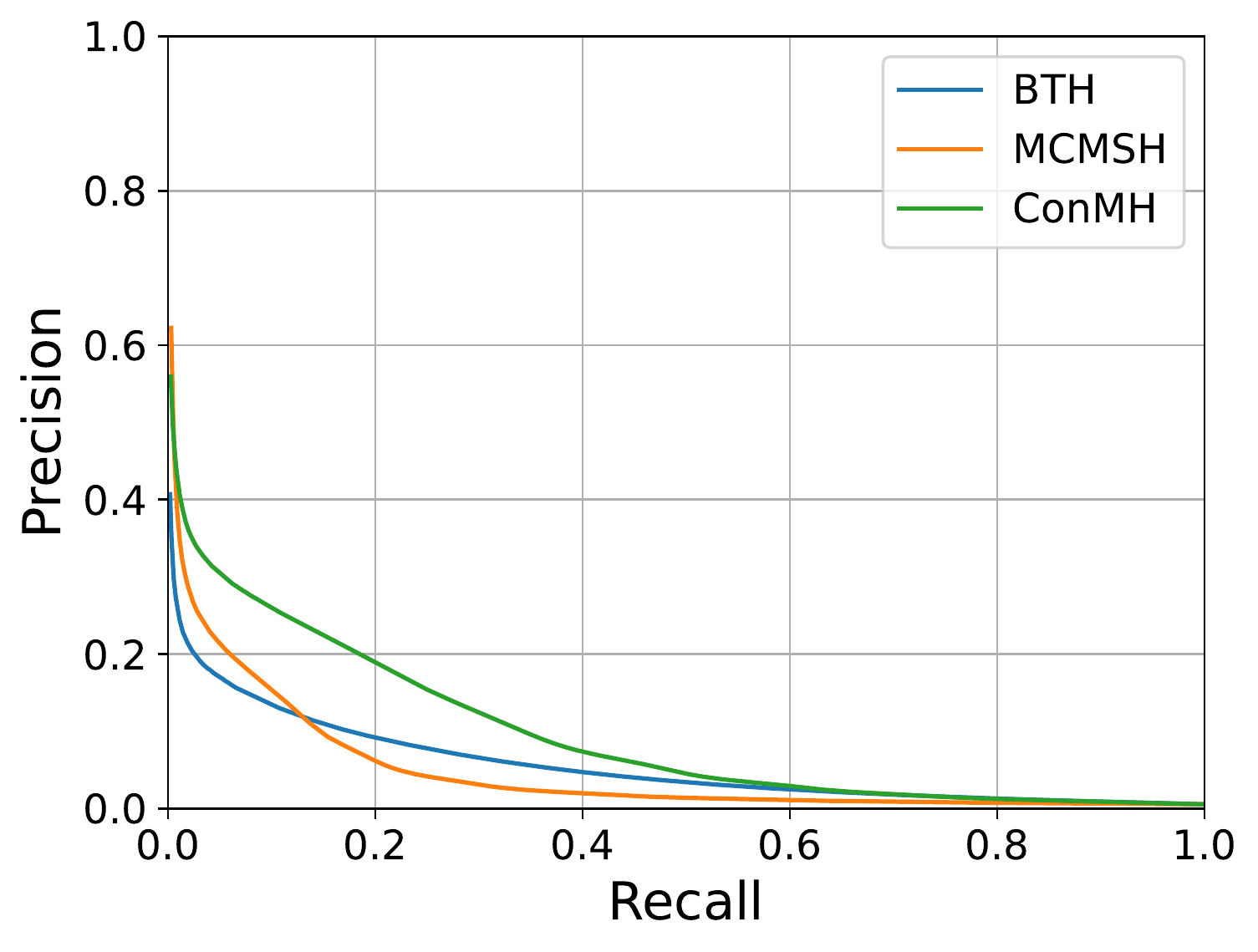}}
  \subfloat[\small{FCVID 32 bits}]{\includegraphics[width = 0.155\textwidth]{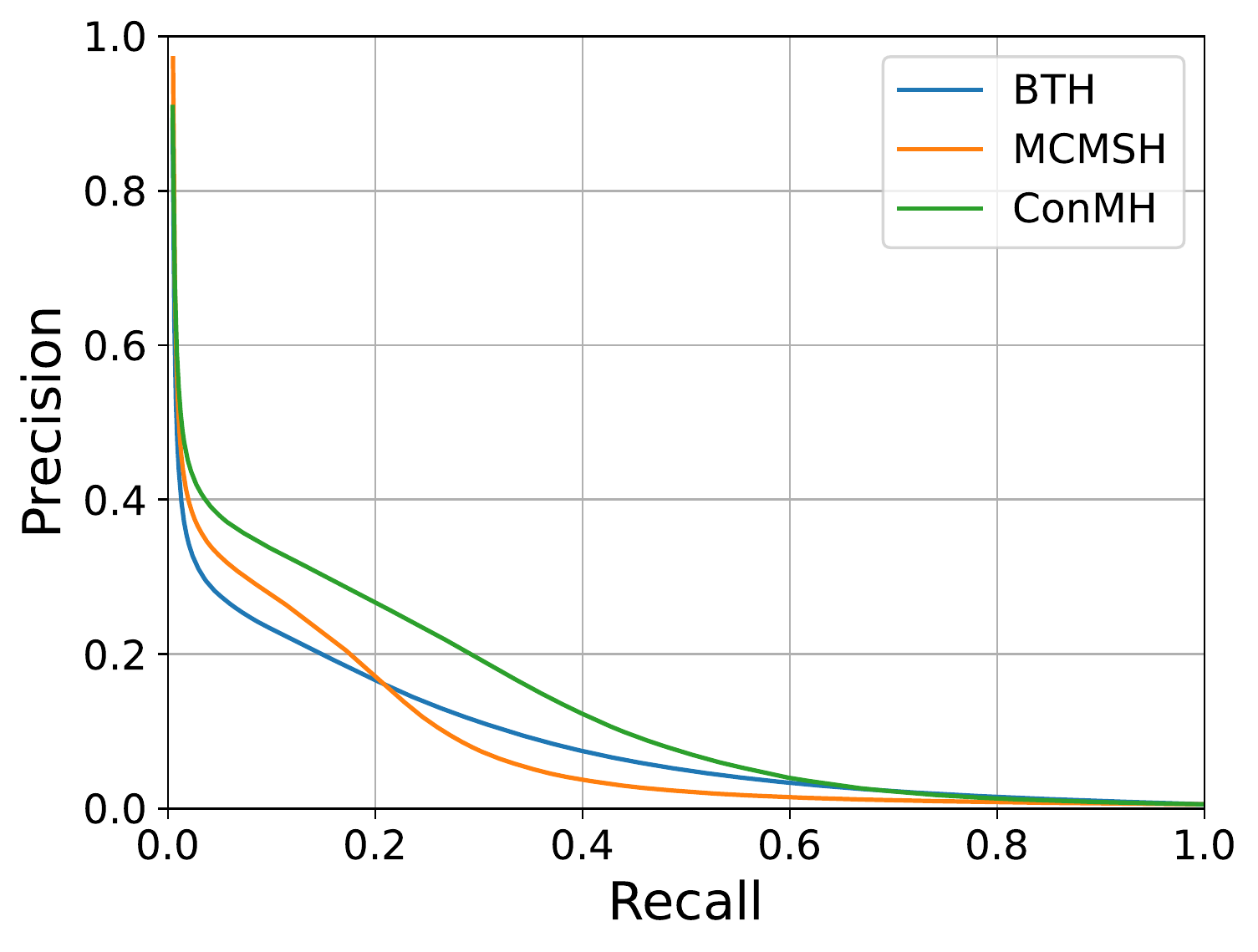}}
  \subfloat[\small{FCVID 64 bits}]{\includegraphics[width = 0.155\textwidth]{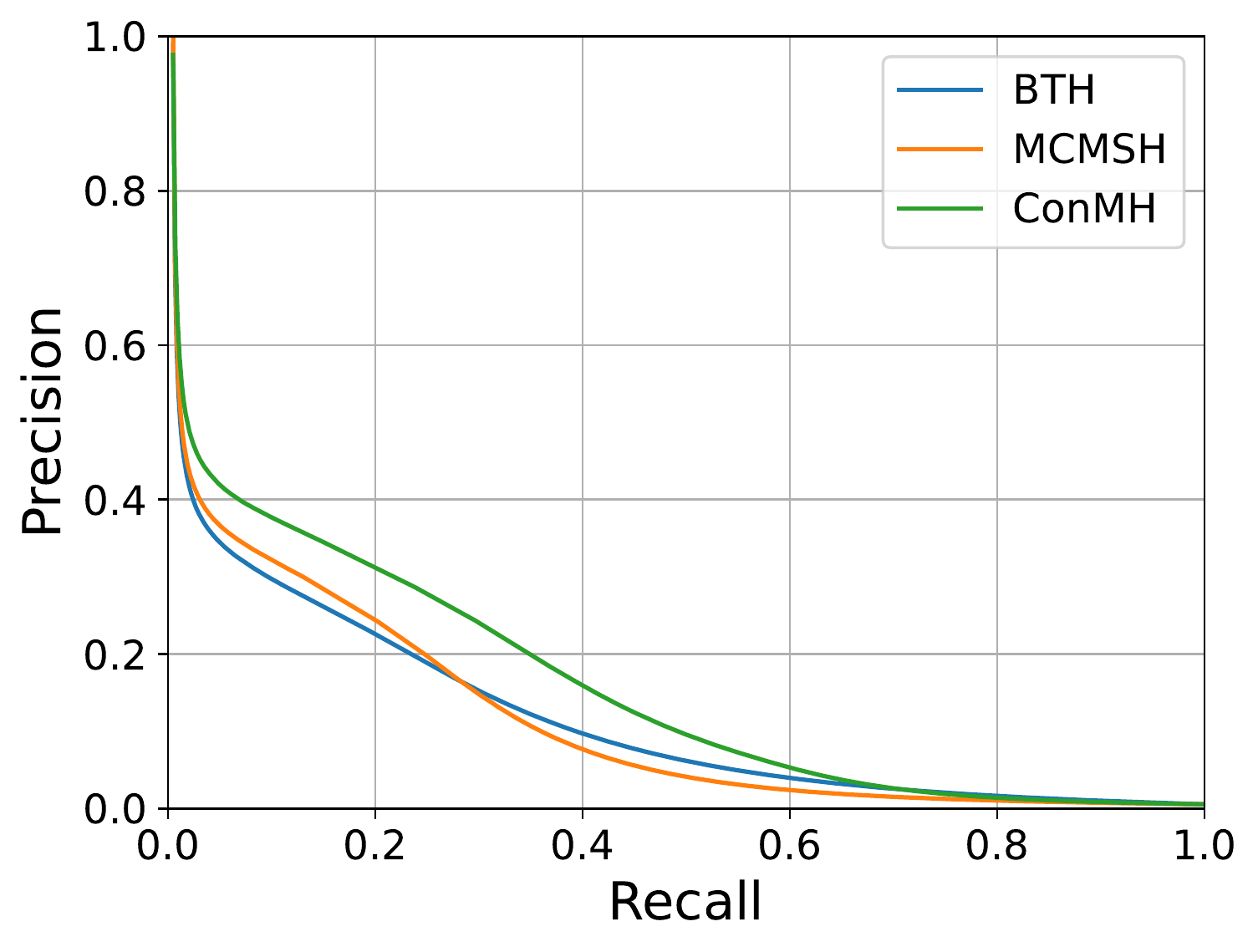}}
  \vspace{-1em}
  \subfloat[\small{Act-Net 16 bits}]{\includegraphics[width = 0.155\textwidth]{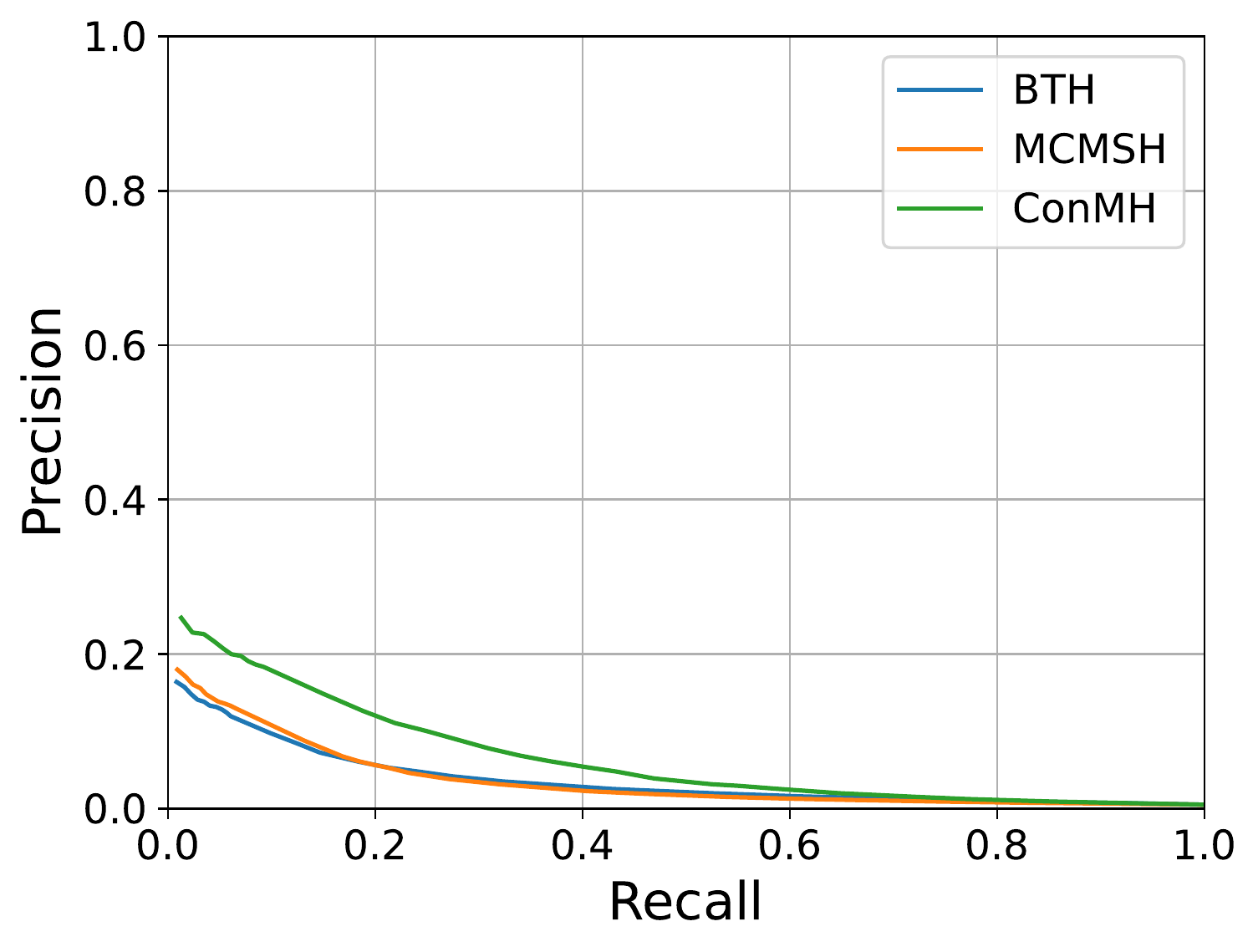}}
  \subfloat[\small{Act-Net 32 bits}]{\includegraphics[width = 0.155\textwidth]{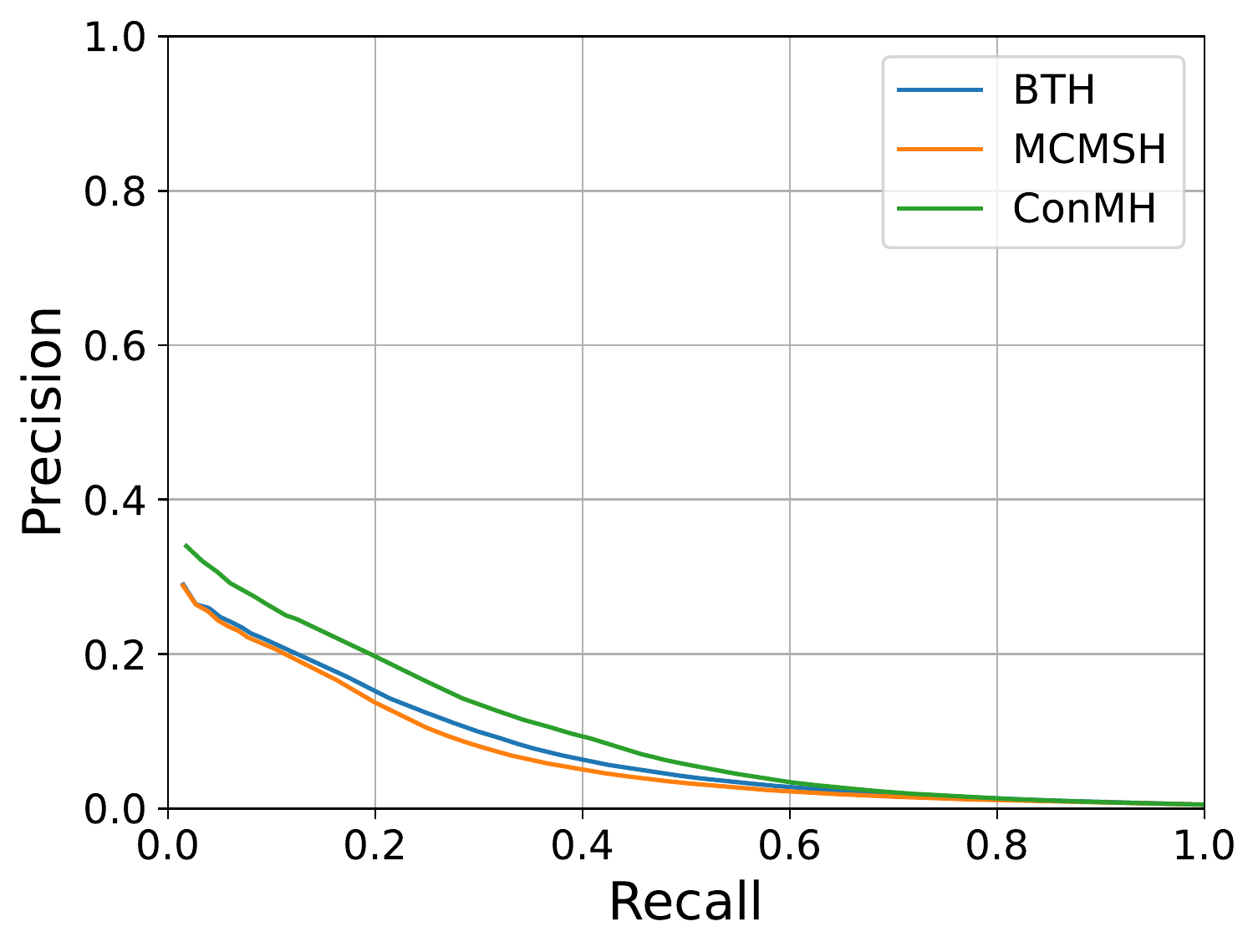}}
  \subfloat[\small{Act-Net 64 bits}]{\includegraphics[width = 0.155\textwidth]{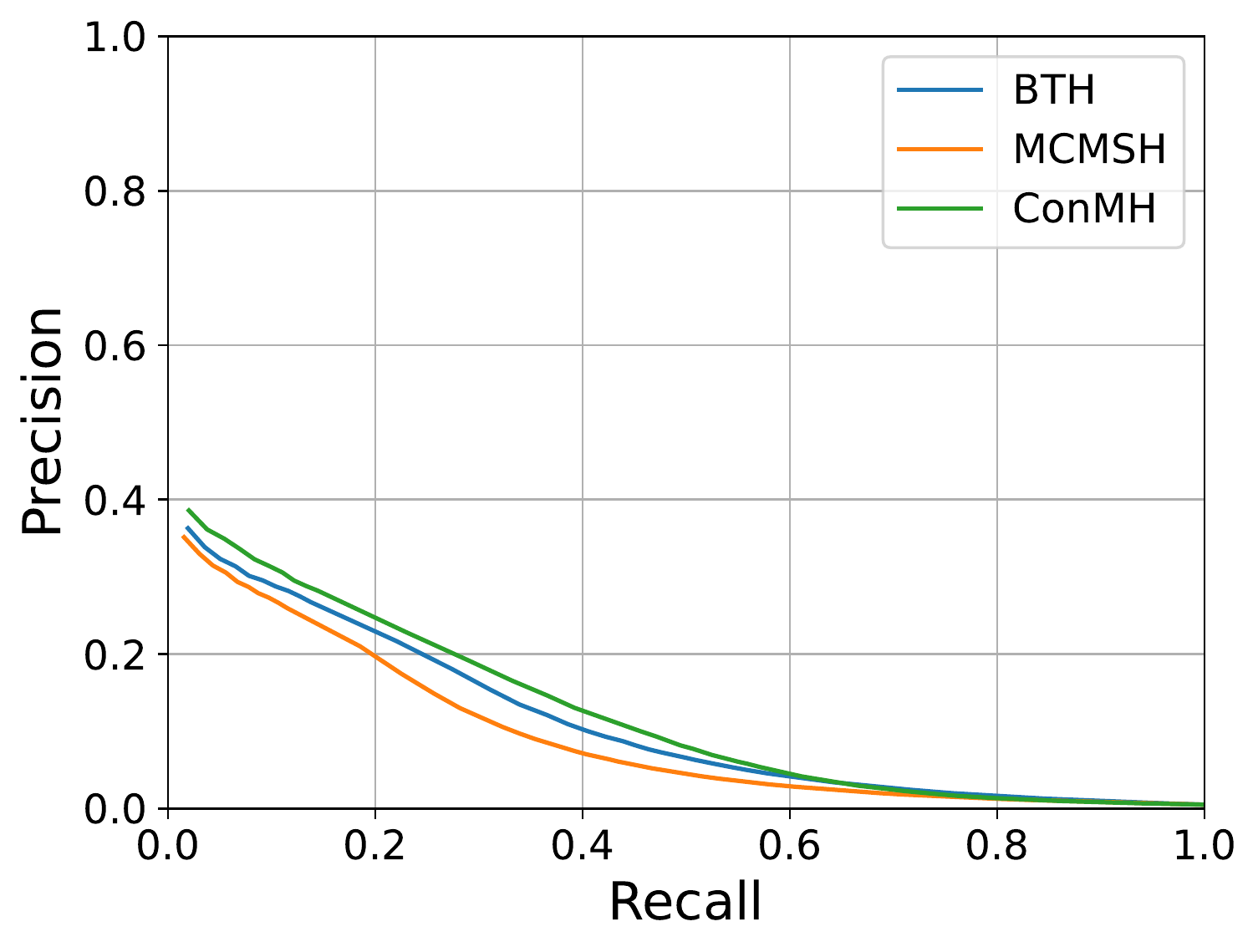}}
  \caption{PR curves of BTH, MCMSH and ConMH on FCVID and ActivityNet.}
  \label{prpr}
\end{figure}

        \textbf{Precision-Recall Curves:} The precision-recall curves of BTH, MCMSH and ConMH on FCVID and ActivityNet are shown in Figure \ref{prpr}. As can be seen, compared with BTH and MCMSH, our ConMH achieves higher precision at the same recall rate with all code lengths. 

        \textbf{Cross-dataset Evaluation Comparisons:} To investigate the generalization of our ConMH, we evaluate cross-dataset retrieval performance among different methods in this subsection. In detail, we train various methods on FCVID and test them on YFCC. We compare the retrieval results with a single-dataset case (both training and testing on YFCC). Table \ref{crossdataset} shows mAP@20 results of various methods with 64 bits in the cross-dataset setting. Although all methods degrade in performance, ConMH still performs the best. Besides, the performance drop of ConMH is negligible, showing the generalization of ConMH across different datasets.

\begin{figure}[h]
  \centering
  \subfloat[MCMSH]{\includegraphics[width = 0.23\textwidth]{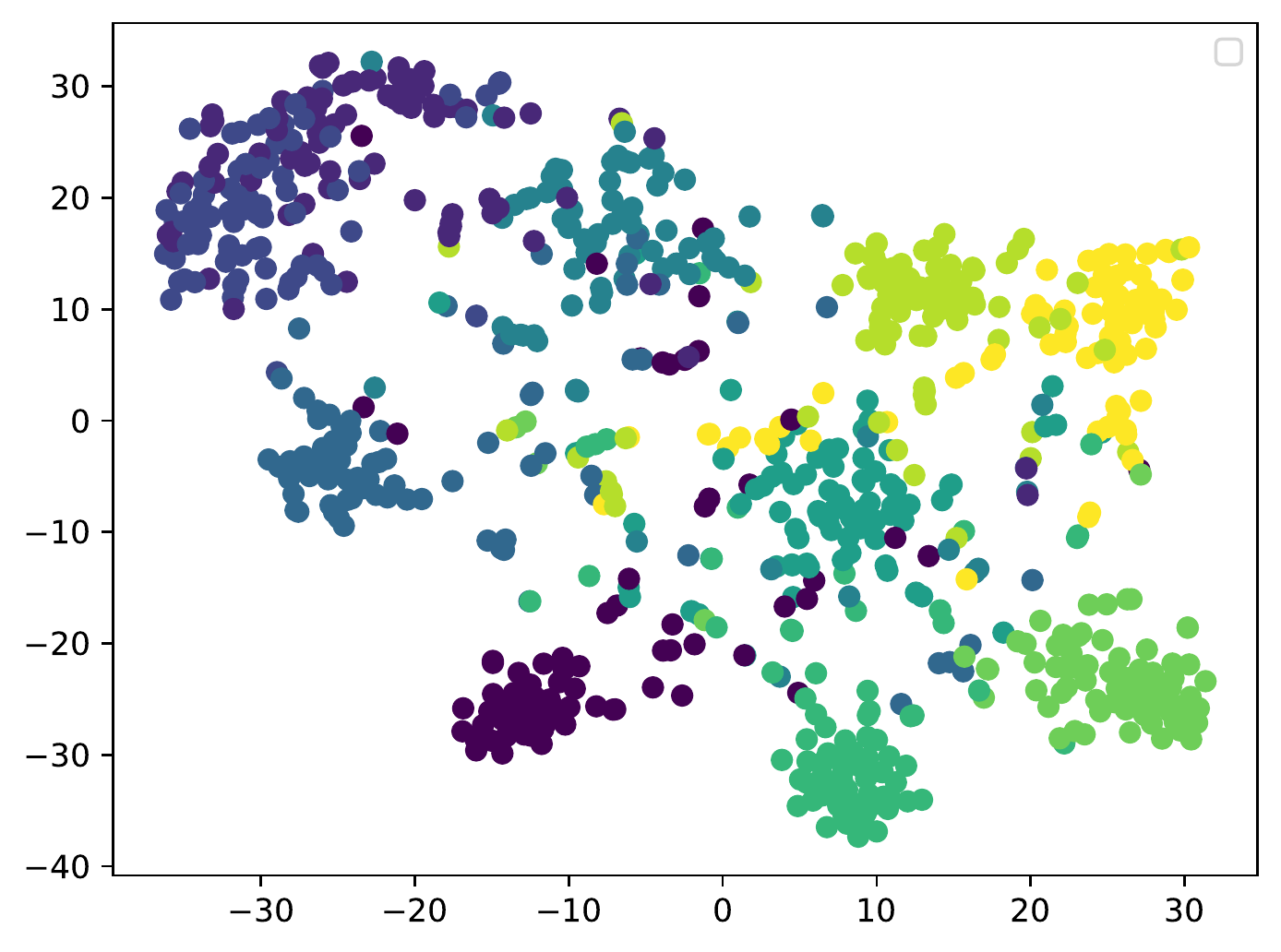}}
  \subfloat[ConMH]{\includegraphics[width = 0.23\textwidth]{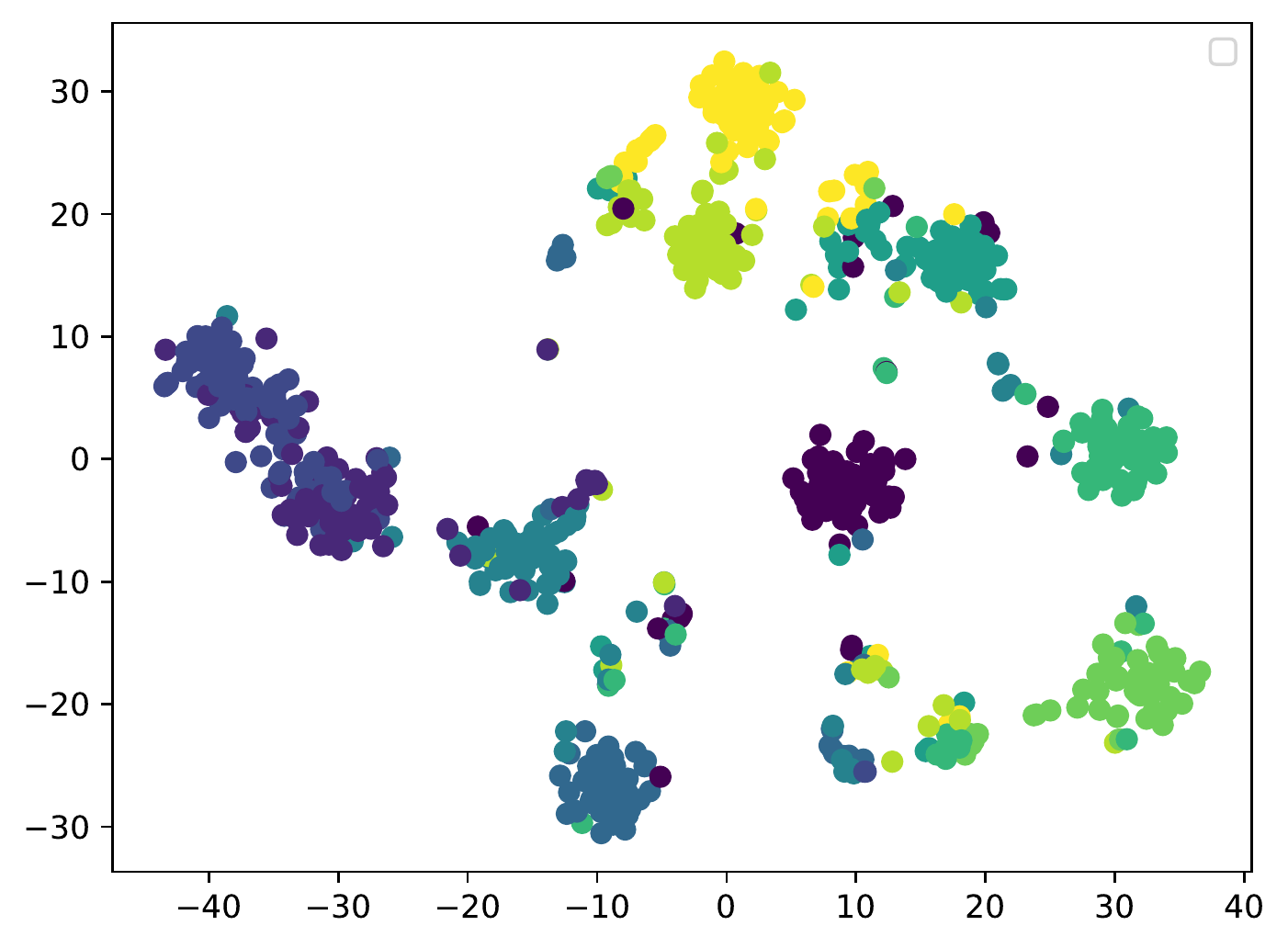}}
  \caption{t-SNE visualizations of MCMSH and ConMH with 64 bits on FCVID-small.}
  \label{tsne}
\end{figure}

    \textbf{Qualitative Results:} We randomly sample 10 categories in FCVID, each with 80 videos, constituting FCVID-small. And we show the t-SNE visualizations of MCMSH and ConMH with 64 bits on FCVID-small.  As shown in Figure \ref{tsne}, hash codes generated by ConMH better distinguish different categories of videos. 

    \subsection{Ablation Study}

        \textbf{Effectiveness of $\mathcal{L}_{recon}$ and $\mathcal{L}_{contra}$:} To analyse the effectiveness of two pretext tasks (\ie, $\mathcal{L}_{recon}$ and $\mathcal{L}_{contra}$) of ConMH, we construct several ConMH variants: (1) ConMH w/o $\mathcal{L}_{contra}$ removes the relational understanding task and only uses  $\mathcal{L}_{recon}$ to train the model. (2) ConMH w/o $\mathcal{L}_{recon}$ removes the reconstruction task and only uses $\mathcal{L}_{contra}$ to train the model. (3) ConMH-norm removes the temporal-masking operation based on ConMH w/o $\mathcal{L}_{contra}$ and reconstructs the full video frames during training. As shown in the first two rows of Table \ref{component}, the results of ConMH w/o $\mathcal{L}_{contra}$ outperform ConMH-norm with all code lengths, showing that the model can better understand the semantic information of video from our highly temporal-masking reconstruction task. Besides, from Table \ref{component}, training with only $\mathcal{L}_{recon}$ or $\mathcal{L}_{contra}$ both lead to the reduction of model performance compared with ConMH. With only  $\mathcal{L}_{recon}$, although the model can understand the semantic information of video, it ignores the similarity relationship between videos which is disadvantageous for the retrieval task. And with only $\mathcal{L}_{contra}$, the model cannot understand the video content well. Therefore, taking care of both the reconstruction and discrimination tasks is critical for good retrieval performance.

        Furthermore, we explain the effectiveness of hashing-oriented learning objectives in ConMH on hash code learning. Most existing methods perform general feature learning (GFL) first, making the hidden representations discriminative. Specifically, they use the hidden representations to complete pretext tasks and binarize them directly to obtain the hash codes. Besides, they introduce a quantization loss to reduce the quantization error from the binarization. Such an approach is hard to balance feature learning and quantization error reduction. However, we directly use the generated hash codes to complete the reconstruction and discriminative tasks, which enables the hash codes to retain more information. We have also tried general feature learning approach to use hidden representations to complete the pretext tasks and introduce a quantization loss, denoted as ConMH w/ GFL. As shown in Table \ref{component}, the hashing-oriented approach is better.

\begin{table}[h]
\centering
    \scalebox{0.58}{
    \begin{tabular}{c|c|c|c|c|c|c|c|c|c}
    \hline
    \multirow{2}{*}{Method} 
    & \multicolumn{3}{c}{16 bits}  \vline   
    & \multicolumn{3}{c}{32 bits}  \vline
    & \multicolumn{3}{c}{64 bits}  \\
    \cline{2-10}
    & K=5 &K=10 &K=20 &K=5 &K=10 &K=20 &K=5 &K=10 &K=20 \\
    \hline
    ConMH-norm & 0.242& 0.156 & 0.103 &
    0.341 &  0.224  & 0.152 & 0.405  & 0.286  & 0.206 \\
    ConMH w/o $\mathcal{L}_{contra}$ & 0.287& 0.208 & 0.156 &
    0.411 &  0.297  & 0.220 & 0.469  & 0.356  & 0.275 \\
    ConMH w/o $\mathcal{L}_{recon}$ & 0.262 & 0.227 & 0.202 &
    0.335  & 0.287  & 0.257 & 0.332  & 0.284  & 0.254 \\
    ConMH w/ GFL & 0.274 & 0.228 & 0.193 &
    0.393  & 0.313  & 0.259 & 0.465  & 0.368  & 0.304 \\
    ConMH &\textbf{0.350} & \textbf{0.293} & \textbf{0.252} &
    \textbf{0.476} &  \textbf{0.390}  & \textbf{0.332} & \textbf{0.524} & 
    \textbf{0.433} &  \textbf{0.373} \\
    \hline
    \hline
    \end{tabular}
    }
    \caption{mAP@K results with different pretext tasks of ConMH on FCVID with 16, 32 and 64 bits.}
      \label{component}
\end{table}

\begin{table}[h]
  
  \centering
  \scalebox{0.68}{
  \begin{tabular}{c|c|c|c|c|c|c|c}
    \hline
    Method & Backbone & K=5 & K=20& K=40& K=60& K=80& K=100 \\
    \hline
    BTH & VGG-16 & 0.475 & 0.310 & 0.260 & 0.235 &  0.216  & 0.202 \\
    % \hline
    MCMSH & VGG-16 & 0.494 & 0.335 & 0.288 & 0.263 &  0.245  & 0.230 \\
    % \hline
    ConMH & VGG-16 & \textbf{0.524} & \textbf{0.373} & \textbf{0.326} & \textbf{0.301} &  \textbf{0.283}  & \textbf{0.267} \\
    \hline
    % \hline
    BTH & Swin Transformer & 0.627 & 0.490 & 0.443 & 0.415 &  0.394  & 0.375 \\
    % \hline
    MCMSH & Swin Transformer & 0.665 & 0.541 & 0.497 & 0.471 &  0.450  & 0.430 \\
    % \hline
    ConMH & Swin Transformer &\textbf{0.675} & \textbf{0.567} & \textbf{0.530} &
    \textbf{0.507} &  \textbf{0.489}  & \textbf{0.472} \\
    \hline
    \hline
  \end{tabular}}
  \caption{mAP@K results of various methods using different backbones with 64 bits.}
  \label{swin}
\end{table}

        \textbf{Retrieval Performance with Swin Transformer:} Recently, Swin Transformer \cite{liu2021swin} has proved its superiority in image feature extracting, compared with the CNN model. To investigate how ConMH performs using a stronger feature extraction backbone, we use the ImageNet-pretrained Swin Transformer backbone to replace the CNN backbone. Then we evaluate the performance of different methods under such a stronger setting. As shown in Table \ref{swin}, with a stronger backbone, ConMH still performs better than BTH and MCMSH, proving the scalability of ConMH.

\begin{figure}[h]
\centering
  \subfloat[Impact of different masking ratios on ConMH performance.]{\includegraphics[width = 0.22\textwidth]{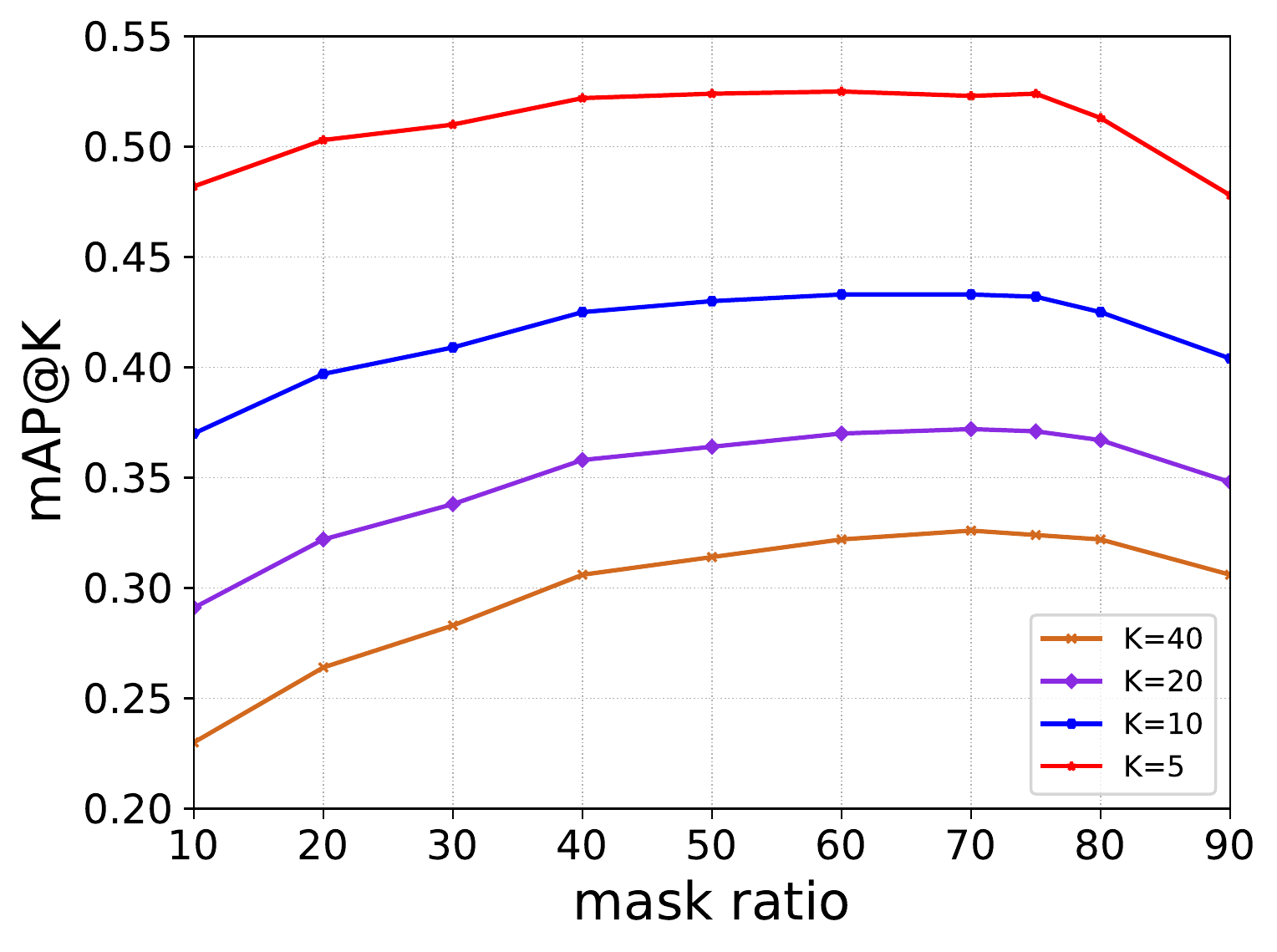}}
  \quad
  \subfloat[Impact of different masking ratios on single-task ConMH performance.]{\includegraphics[width = 0.22\textwidth]{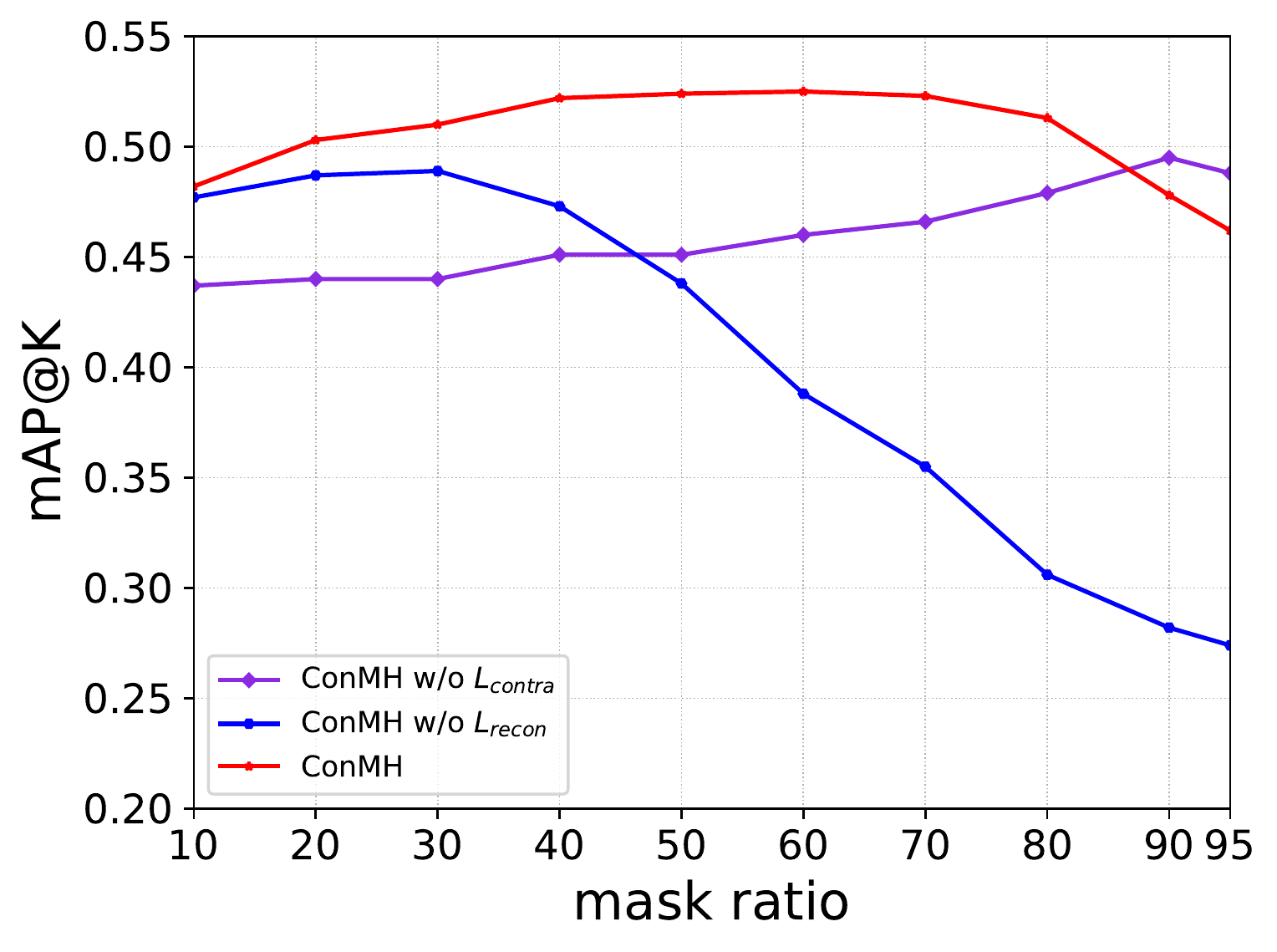}}
  \caption{\textbf{Retrieval performance with different masking ratios on FCVID with 64 bits.} A high masking ratio works well. However, an extremely high masking ratio degrades performance, in which case the model cannot learn useful knowledge from the contrastive task.}
  \label{maskratio}
\end{figure}

        \textbf{Effects of Different Masking Ratios:} We evaluate model performance with different masking ratios on FCVID with 64 bits. As shown in Figure \ref{maskratio} (a), a high masking ratio (\ie, 50\% to 75\%) works well. Both low and extremely high masking ratios will degrade model performance. Furthermore, we study the effect of masking ratios in the case of only a single task. As shown in Figure \ref{maskratio} (b), when only the reconstruction task is employed, a high masking ratio (e.g., 90\%) works better. As for the contrastive task, a moderately lower masking ratio (e.g., 30\%) is preferred. The reasons are two-fold. On the one hand, a high masking ratio will benefit intra-sample semantic understanding via reconstruction task. On the other hand, the large information loss it incurs will increase the difficulty of learning semantically-invariant information between positive samples in the contrastive task and even harm the task convergence. Besides, we find that adopting a compromise masking ratio (e.g., 50\%-75\%) helps to balance and unify two objectives in one learning framework effectively. As we can learn from the above results, the complete version of ConMH outperforms both single-task models when choosing proper masking ratios.

        \textbf{Effects of Different Masking Strategies:} In this subsection, we investigate the impact of different masking strategies. From Table \ref{mask_strategy}, we can learn that the entire frame masking strategy achieves the best performance, while tube-based masking and token-based masking strategies show not very satisfactory results. The reason might be that our method (which follows the standard SSVH settings) intrinsically operates on frame-level granularity. Therefore, when combining our method with the tube-based or token-based masking strategies, the input video frames will suffer from a large proportion of information loss. It results in weak frame-level features and thus adversely reduces the effectiveness of the reconstruction task. Besides, we have to preserve all the frames for the reconstruction task and suffer from larger memory overhead. 

\begin{table}[h]
  \centering
  \scalebox{0.78}{
  \begin{tabular}{c|c|c|c|c|c|c}
    \hline
    Masking Strategy & K=5 & K=20& K=40& K=60& K=80& K=100 \\
    \hline
    tube-based mask & 0.387 & 0.198 & 0.147 &
    0.124 &  0.109  & 0.099 \\
    % \hline
    token-based mask & 0.412 & 0.227 & 0.175 &
    0.151 &  0.135  & 0.129 \\
    % \hline
    entire frame mask &\textbf{0.524} & \textbf{0.373} & \textbf{0.326} &
    \textbf{0.301} &  \textbf{0.283}  & \textbf{0.267} \\
    \hline
    \hline
  \end{tabular}}
  \caption{mAP@K results of different masking strategies on FCVID with 64 bits.}
  \label{mask_strategy}
\end{table}

        \textbf{Effects of Different Sampling Strategies:} In this subsection, we investigate the impact of different sampling strategies as shown in Table \ref{sample_strategy}. In general, both overlapped and non-overlapped sampling strategies behave similarly in terms of performance. It might be due to the information redundancy among successive frames. Besides, the long and short sampling strategy does not exhibit an advantage in our case. We guess the reason might be that the constrained pattern reduces the sampling's diversity (or randomness), which might be less effective on untrimmed videos.

\begin{table}[h]
  \centering
  \scalebox{0.78}{
  \begin{tabular}{c|c|c|c|c|c|c}
    \hline
    Sampling Strategy & K=5 & K=20& K=40& K=60& K=80& K=100 \\
    \hline
    long and short way & 0.509 & 0.355 & 0.308 &
    0.283 &  0.265  & 0.250 \\
    % \hline
    overlapped & \textbf{0.528} & \textbf{0.374} & \textbf{0.327} &
    0.300 &  0.282  & 0.265 \\
    % \hline
    non-overlapped &0.524 & 0.373 & 0.326 &
    \textbf{0.301} &  \textbf{0.283}  & \textbf{0.267} \\
    \hline
    \hline
  \end{tabular}}
  \caption{mAP@K results of different sampling strategies on FCVID with 64 bits.}
  \label{sample_strategy}
\end{table}

    \textbf{Effects of Different Model Sizes:} In this subsection, we explore the impact of different model sizes on FCVID with 64 bits. As shown in the first ten rows of Table \ref{modeltype}, using a small decoder to reconstruct the original videos is better. When the decoder is big (\eg, with 12 depth or 1024 width), the retrieval performance of the model, in turn, will degrade. That is because a small decoder needs more semantic information to complete the reconstruction task, forcing the encoder to generate more informative representations and hash tokens. However, the decoder cannot be too small (\eg, too small width or a simple FC layer), which will make the reconstruction task too difficult to accomplish, leading to poor performance. Although the decoder structure can be chosen arbitrarily because it will be discarded during testing, it still affects the training time. Therefore, we comprehensively consider the training time and model performance, then set the depth and width of the decoder as 2 and 192, respectively. Besides, as shown in the last three rows of Table \ref{modeltype}, when the encoder size becomes larger, the performance of the model is correspondingly improved, demonstrating that ConMH is a scalable framework for self-supervised video hashing, which still works well even with a huge model structure. However, large model structures make inference time long, violating the real-time requirement of video hashing. So we choose ViT-small as our defaulting encoder structure.

\begin{table}[h]
  \centering
  \scalebox{0.78}{
  \begin{tabular}{c c|c|c|c|c|c|c|c}
    \hline
    \multicolumn{2}{c}{\multirow{2}{*}{Encoder}} \vline & \multicolumn{2}{c}{Decoder} \vline &
    \multirow{2}{*}{K=20} &\multirow{2}{*}{K=40} &
    \multirow{2}{*}{K=60} &\multirow{2}{*}{K=80} & 
    \multirow{2}{*}{K=100} \\
    \cline{3-4}
    & &  depth & width & & & & & \\
    \hline
    \multicolumn{2}{c}{\multirow{10}{*}{\textbf{ViT-small}}} \vline & 0 & 
    \textbf{\multirow{6}{*}{192}} &
    0.254 & 0.232 & 0.221 & 0.212 & 0.205 \\
    % \hline
    & & 1 & & 0.372 & 0.326 & 0.301 & 0.282 & 0.266 \\
    % \hline
    & & \textbf{2} & & \textbf{0.373} & \textbf{0.326} & \textbf{0.301} & \textbf{0.283} & \textbf{0.267} \\
    % \hline
    & & 4 & & 0.371 & 0.324 & 0.299 & 0.280 & 0.264 \\
    % \hline
    & & 8 & & 0.371 & 0.324 & 0.299 & 0.280 & 0.264 \\
    % \hline
    & & 12 & & 0.366 & 0.320 & 0.295 & 0.276 & 0.260 \\
    \cline{3-9}
    & & \multirow{4}{*}{2} & 64 & 0.358 & 0.314 & 0.291 & 0.273 & 0.258 \\
    & & & 256 & 0.373 & 0.326 & 0.300 & 0.280 & 0.264 \\
    & & & 512 & 0.372 & 0.324 & 0.298 & 0.278 & 0.262 \\
    & & & 1024 & 0.370 & 0.322 & 0.296 & 0.277 & 0.260 \\
    \hline
    \multicolumn{2}{c}{\multirow{1}{*}{ViT-mini}} \vline & \multirow{3}{*}{2} & 
    \multirow{3}{*}{192} & 0.324 & 0.274 & 0.248 & 0.229 & 0.214 \\
    \multicolumn{2}{c}{\multirow{1}{*}{ViT-base}} \vline &  &  & 0.397 & 0.349 & 0.323 & 0.303 & 0.285 \\
    \multicolumn{2}{c}{\multirow{1}{*}{ViT-large}} \vline &  &  & 0.401 & 0.352 & 0.325 & 0.305 & 0.287 \\
    \hline
    \hline
  \end{tabular}}
  \caption{64 bits ConMH mAP@K results with different encoder-decoder sizes on FCVID. 
  The default settings for ConMH are bolded. Decoder depth is 0 means that we use a simple FC layer to reconstruct the original frames.}
  \label{modeltype}
\end{table}

\section{Conclusions}

    In this paper, we propose a simple yet effective contrastive masked autoencoders framework called ConMH for self-supervised video hashing. ConMH uses an asymmetric encoder-decoder framework to better understand the video's visual semantic content. The encoder takes highly temporal-masked frames to generate hash codes, and the decoder reconstructs the masked video frames by utilizing generated hash tokens and masked tokens. We find that a high masking ratio is beneficial. Besides, ConMH combines a discriminative task to fully exploit the similarity relationship between videos in a single stage. Extensive experiments on three widely used benchmark datasets demonstrate the superiority of our ConMH.

\section{Acknowledge}
This work is supported in part by the National Natural Science Foundation of China under grant 62171248, the PCNL KEY project (PCL2021A07), the Guangdong Basic and Applied Basic Research Foundation under grant 2021A1515110066, and the GXWD 20220811172936001, and Shenzhen Science and Technology Program under Grant JCYJ20220818101012025.

This work was partly conducted based on the undergraduate graduation thesis in Huazhong University of Science and Technology, Wuhan, China.

\bibliography{aaai23}

\begin{thebibliography}{39}
\providecommand{\natexlab}[1]{#1}

\bibitem[{Assran et~al.(2022)Assran, Caron, Misra, Bojanowski, Bordes, Vincent,
  Joulin, Rabbat, and Ballas}]{assran2022masked}
Assran, M.; Caron, M.; Misra, I.; Bojanowski, P.; Bordes, F.; Vincent, P.;
  Joulin, A.; Rabbat, M.; and Ballas, N. 2022.
\newblock Masked siamese networks for label-efficient learning.
\newblock In \emph{European Conference on Computer Vision}, 456--473. Springer.

\bibitem[{Bao, Dong, and Wei(2021)}]{bao2021beit}
Bao, H.; Dong, L.; and Wei, F. 2021.
\newblock Beit: Bert pre-training of image transformers.
\newblock \emph{arXiv preprint arXiv:2106.08254}.

\bibitem[{Caba~Heilbron et~al.(2015)Caba~Heilbron, Escorcia, Ghanem, and
  Carlos~Niebles}]{caba2015activitynet}
Caba~Heilbron, F.; Escorcia, V.; Ghanem, B.; and Carlos~Niebles, J. 2015.
\newblock Activitynet: A large-scale video benchmark for human activity
  understanding.
\newblock In \emph{Proceedings of the ieee conference on computer vision and
  pattern recognition}, 961--970.

\bibitem[{Caron et~al.(2020)Caron, Misra, Mairal, Goyal, Bojanowski, and
  Joulin}]{caron2020unsupervised}
Caron, M.; Misra, I.; Mairal, J.; Goyal, P.; Bojanowski, P.; and Joulin, A.
  2020.
\newblock Unsupervised learning of visual features by contrasting cluster
  assignments.
\newblock \emph{Advances in Neural Information Processing Systems}, 33:
  9912--9924.

\bibitem[{Chen et~al.(2020)Chen, Kornblith, Norouzi, and
  Hinton}]{chen2020simple}
Chen, T.; Kornblith, S.; Norouzi, M.; and Hinton, G. 2020.
\newblock A simple framework for contrastive learning of visual
  representations.
\newblock In \emph{International conference on machine learning}, 1597--1607.
  PMLR.

\bibitem[{Chen et~al.(2022)Chen, Ding, Wang, Xin, Mo, Wang, Han, Luo, Zeng, and
  Wang}]{chen2022context}
Chen, X.; Ding, M.; Wang, X.; Xin, Y.; Mo, S.; Wang, Y.; Han, S.; Luo, P.;
  Zeng, G.; and Wang, J. 2022.
\newblock Context autoencoder for self-supervised representation learning.
\newblock \emph{arXiv preprint arXiv:2202.03026}.

\bibitem[{Chen and He(2021)}]{chen2021exploring}
Chen, X.; and He, K. 2021.
\newblock Exploring simple siamese representation learning.
\newblock In \emph{Proceedings of the IEEE/CVF Conference on Computer Vision
  and Pattern Recognition}, 15750--15758.

\bibitem[{Chuang et~al.(2020)Chuang, Robinson, Lin, Torralba, and
  Jegelka}]{chuang2020debiased}
Chuang, C.-Y.; Robinson, J.; Lin, Y.-C.; Torralba, A.; and Jegelka, S. 2020.
\newblock Debiased contrastive learning.
\newblock \emph{Advances in neural information processing systems}, 33:
  8765--8775.

\bibitem[{Cui et~al.(2021)Cui, Zhu, Li, Cheng, and Zhang}]{cui2021two}
Cui, H.; Zhu, L.; Li, J.; Cheng, Z.; and Zhang, Z. 2021.
\newblock Two-pronged Strategy: Lightweight Augmented Graph Network Hashing for
  Scalable Image Retrieval.
\newblock In \emph{Proceedings of the 29th ACM International Conference on
  Multimedia}, 1432--1440.

\bibitem[{Dong et~al.(2021)Dong, Bao, Zhang, Chen, Zhang, Yuan, Chen, Wen, and
  Yu}]{dong2021peco}
Dong, X.; Bao, J.; Zhang, T.; Chen, D.; Zhang, W.; Yuan, L.; Chen, D.; Wen, F.;
  and Yu, N. 2021.
\newblock PeCo: Perceptual Codebook for BERT Pre-training of Vision
  Transformers.
\newblock \emph{arXiv preprint arXiv:2111.12710}.

\bibitem[{Dosovitskiy et~al.(2020)Dosovitskiy, Beyer, Kolesnikov, Weissenborn,
  Zhai, Unterthiner, Dehghani, Minderer, Heigold, Gelly
  et~al.}]{dosovitskiy2020image}
Dosovitskiy, A.; Beyer, L.; Kolesnikov, A.; Weissenborn, D.; Zhai, X.;
  Unterthiner, T.; Dehghani, M.; Minderer, M.; Heigold, G.; Gelly, S.; et~al.
  2020.
\newblock An image is worth 16x16 words: Transformers for image recognition at
  scale.
\newblock \emph{arXiv preprint arXiv:2010.11929}.

\bibitem[{Erin~Liong et~al.(2015)Erin~Liong, Lu, Wang, Moulin, and
  Zhou}]{erin2015deep}
Erin~Liong, V.; Lu, J.; Wang, G.; Moulin, P.; and Zhou, J. 2015.
\newblock Deep hashing for compact binary codes learning.
\newblock In \emph{Proceedings of the IEEE conference on computer vision and
  pattern recognition}, 2475--2483.

\bibitem[{Gao et~al.(2015)Gao, Song, Zou, Zhang, and Shao}]{gao2015scalable}
Gao, L.; Song, J.; Zou, F.; Zhang, D.; and Shao, J. 2015.
\newblock Scalable multimedia retrieval by deep learning hashing with relative
  similarity learning.
\newblock In \emph{Proceedings of the 23rd ACM international conference on
  Multimedia}, 903--906.

\bibitem[{Gong et~al.(2012)Gong, Lazebnik, Gordo, and
  Perronnin}]{gong2012iterative}
Gong, Y.; Lazebnik, S.; Gordo, A.; and Perronnin, F. 2012.
\newblock Iterative quantization: A procrustean approach to learning binary
  codes for large-scale image retrieval.
\newblock \emph{IEEE transactions on pattern analysis and machine
  intelligence}, 35(12): 2916--2929.

\bibitem[{Grill et~al.(2020)Grill, Strub, Altch{\'e}, Tallec, Richemond,
  Buchatskaya, Doersch, Avila~Pires, Guo, Gheshlaghi~Azar
  et~al.}]{grill2020bootstrap}
Grill, J.-B.; Strub, F.; Altch{\'e}, F.; Tallec, C.; Richemond, P.;
  Buchatskaya, E.; Doersch, C.; Avila~Pires, B.; Guo, Z.; Gheshlaghi~Azar, M.;
  et~al. 2020.
\newblock Bootstrap your own latent-a new approach to self-supervised learning.
\newblock \emph{Advances in Neural Information Processing Systems}, 33:
  21271--21284.

\bibitem[{Hao et~al.(2022)Hao, Duan, Zhang, Zhu, Zhou, and
  He}]{hao2022unsupervised}
Hao, Y.; Duan, J.; Zhang, H.; Zhu, B.; Zhou, P.; and He, X. 2022.
\newblock Unsupervised Video Hashing with Multi-granularity Contextualization
  and Multi-structure Preservation.
\newblock In \emph{ACM International Conference on Multimedia (MM’22)}.

\bibitem[{He et~al.(2022)He, Chen, Xie, Li, Doll{\'a}r, and
  Girshick}]{he2021masked}
He, K.; Chen, X.; Xie, S.; Li, Y.; Doll{\'a}r, P.; and Girshick, R. 2022.
\newblock Masked autoencoders are scalable vision learners.
\newblock In \emph{Proceedings of the IEEE/CVF Conference on Computer Vision
  and Pattern Recognition}, 16000--16009.

\bibitem[{He et~al.(2020)He, Fan, Wu, Xie, and Girshick}]{he2020momentum}
He, K.; Fan, H.; Wu, Y.; Xie, S.; and Girshick, R. 2020.
\newblock Momentum contrast for unsupervised visual representation learning.
\newblock In \emph{Proceedings of the IEEE/CVF conference on computer vision
  and pattern recognition}, 9729--9738.

\bibitem[{He et~al.(2016)He, Zhang, Ren, and Sun}]{he2016deep}
He, K.; Zhang, X.; Ren, S.; and Sun, J. 2016.
\newblock Deep residual learning for image recognition.
\newblock In \emph{Proceedings of the IEEE conference on computer vision and
  pattern recognition}, 770--778.

\bibitem[{Jiang et~al.(2017)Jiang, Wu, Wang, Xue, and
  Chang}]{jiang2017exploiting}
Jiang, Y.-G.; Wu, Z.; Wang, J.; Xue, X.; and Chang, S.-F. 2017.
\newblock Exploiting feature and class relationships in video categorization
  with regularized deep neural networks.
\newblock \emph{IEEE transactions on pattern analysis and machine
  intelligence}, 40(2): 352--364.

\bibitem[{Kingma and Ba(2014)}]{kingma2014adam}
Kingma, D.~P.; and Ba, J. 2014.
\newblock Adam: A method for stochastic optimization.
\newblock \emph{arXiv preprint arXiv:1412.6980}.

\bibitem[{Li et~al.(2017)Li, Yang, Cao, and Huang}]{li2017jointly}
Li, C.; Yang, Y.; Cao, J.; and Huang, Z. 2017.
\newblock Jointly modeling static visual appearance and temporal pattern for
  unsupervised video hashing.
\newblock In \emph{Proceedings of the 2017 ACM on Conference on Information and
  Knowledge Management}, 9--17.

\bibitem[{Li et~al.(2019)Li, Chen, Lu, Li, and Zhou}]{li2019neighborhood}
Li, S.; Chen, Z.; Lu, J.; Li, X.; and Zhou, J. 2019.
\newblock Neighborhood preserving hashing for scalable video retrieval.
\newblock In \emph{Proceedings of the IEEE/CVF International Conference on
  Computer Vision}, 8212--8221.

\bibitem[{Li et~al.(2021)Li, Li, Lu, and Zhou}]{li2021self}
Li, S.; Li, X.; Lu, J.; and Zhou, J. 2021.
\newblock Self-supervised video hashing via bidirectional transformers.
\newblock In \emph{Proceedings of the IEEE/CVF Conference on Computer Vision
  and Pattern Recognition}, 13549--13558.

\bibitem[{Liu et~al.(2021)Liu, Lin, Cao, Hu, Wei, Zhang, Lin, and
  Guo}]{liu2021swin}
Liu, Z.; Lin, Y.; Cao, Y.; Hu, H.; Wei, Y.; Zhang, Z.; Lin, S.; and Guo, B.
  2021.
\newblock Swin transformer: Hierarchical vision transformer using shifted
  windows.
\newblock In \emph{Proceedings of the IEEE/CVF International Conference on
  Computer Vision}, 10012--10022.

\bibitem[{Lu et~al.(2021)Lu, Wang, Zeng, Chen, Wu, and Xia}]{SwinFGHash}
Lu, D.; Wang, J.; Zeng, Z.; Chen, B.; Wu, S.; and Xia, S.-T. 2021.
\newblock SwinFGHash: Fine-grained Image Retrieval via Transformer-based
  Hashing Network.
\newblock In \emph{32nd British Machine Vision Conference}.

\bibitem[{Luo et~al.(2018)Luo, Nie, He, Wu, Chen, and Xu}]{luo2018fast}
Luo, X.; Nie, L.; He, X.; Wu, Y.; Chen, Z.-D.; and Xu, X.-S. 2018.
\newblock Fast scalable supervised hashing.
\newblock In \emph{The 41st international ACM SIGIR conference on research \&
  development in information retrieval}, 735--744.

\bibitem[{Russakovsky et~al.(2015)Russakovsky, Deng, Su, Krause, Satheesh, Ma,
  Huang, Karpathy, Khosla, Bernstein et~al.}]{russakovsky2015imagenet}
Russakovsky, O.; Deng, J.; Su, H.; Krause, J.; Satheesh, S.; Ma, S.; Huang, Z.;
  Karpathy, A.; Khosla, A.; Bernstein, M.; et~al. 2015.
\newblock Imagenet large scale visual recognition challenge.
\newblock \emph{International journal of computer vision}, 115(3): 211--252.

\bibitem[{Simonyan and Zisserman(2014)}]{simonyan2014very}
Simonyan, K.; and Zisserman, A. 2014.
\newblock Very deep convolutional networks for large-scale image recognition.
\newblock \emph{arXiv preprint arXiv:1409.1556}.

\bibitem[{Song et~al.(2011)Song, Yang, Huang, Shen, and
  Hong}]{song2011multiple}
Song, J.; Yang, Y.; Huang, Z.; Shen, H.~T.; and Hong, R. 2011.
\newblock Multiple feature hashing for real-time large scale near-duplicate
  video retrieval.
\newblock In \emph{Proceedings of the 19th ACM international conference on
  Multimedia}, 423--432.

\bibitem[{Song et~al.(2018)Song, Zhang, Li, Gao, Wang, and Hong}]{song2018self}
Song, J.; Zhang, H.; Li, X.; Gao, L.; Wang, M.; and Hong, R. 2018.
\newblock Self-supervised video hashing with hierarchical binary auto-encoder.
\newblock \emph{IEEE Transactions on Image Processing}, 27(7): 3210--3221.

\bibitem[{Thomee et~al.(2015)Thomee, Shamma, Friedland, Elizalde, Ni, Poland,
  Borth, and Li}]{thomee2015new}
Thomee, B.; Shamma, D.~A.; Friedland, G.; Elizalde, B.; Ni, K.; Poland, D.;
  Borth, D.; and Li, L.-J. 2015.
\newblock The new data and new challenges in multimedia research.
\newblock \emph{arXiv preprint arXiv:1503.01817}, 1(8).

\bibitem[{Vaswani et~al.(2017)Vaswani, Shazeer, Parmar, Uszkoreit, Jones,
  Gomez, Kaiser, and Polosukhin}]{vaswani2017attention}
Vaswani, A.; Shazeer, N.; Parmar, N.; Uszkoreit, J.; Jones, L.; Gomez, A.~N.;
  Kaiser, {\L}.; and Polosukhin, I. 2017.
\newblock Attention is all you need.
\newblock \emph{Advances in neural information processing systems}, 30.

\bibitem[{Wang et~al.(2022)Wang, Zeng, Chen, Wang, Liao, Li, Wang, and
  Xia}]{HuggingHash}
Wang, J.; Zeng, Z.; Chen, B.; Wang, Y.; Liao, D.; Li, G.; Wang, Y.; and Xia,
  S.-T. 2022.
\newblock Hugs Are Better Than Handshakes: Unsupervised Cross-Modal Transformer
  Hashing with Multi-granularity Alignment.
\newblock In \emph{33nd British Machine Vision Conference}.

\bibitem[{Weiss, Torralba, and Fergus(2008)}]{weiss2008spectral}
Weiss, Y.; Torralba, A.; and Fergus, R. 2008.
\newblock Spectral hashing.
\newblock \emph{Advances in neural information processing systems}, 21.

\bibitem[{Wu et~al.(2017)Wu, Liu, Guo, Ding, Han, Shen, and
  Shao}]{wu2017unsupervised}
Wu, G.; Liu, L.; Guo, Y.; Ding, G.; Han, J.; Shen, J.; and Shao, L. 2017.
\newblock Unsupervised deep video hashing with balanced rotation.
\newblock IJCAI.

\bibitem[{Xie et~al.(2022)Xie, Zhang, Cao, Lin, Bao, Yao, Dai, and
  Hu}]{xie2021simmim}
Xie, Z.; Zhang, Z.; Cao, Y.; Lin, Y.; Bao, J.; Yao, Z.; Dai, Q.; and Hu, H.
  2022.
\newblock Simmim: A simple framework for masked image modeling.
\newblock In \emph{Proceedings of the IEEE/CVF Conference on Computer Vision
  and Pattern Recognition}, 9653--9663.

\bibitem[{Zeng et~al.(2022)Zeng, Wang, Chen, Wang, and Xia}]{MAGRH}
Zeng, Z.; Wang, J.; Chen, B.; Wang, Y.; and Xia, S.-T. 2022.
\newblock Motion-Aware Graph Reasoning Hashing for Self-supervised Video
  Retrieval.
\newblock In \emph{33nd British Machine Vision Conference}.

\bibitem[{Zhang et~al.(2016)Zhang, Wang, Hong, and Chua}]{zhang2016play}
Zhang, H.; Wang, M.; Hong, R.; and Chua, T.-S. 2016.
\newblock Play and rewind: Optimizing binary representations of videos by
  self-supervised temporal hashing.
\newblock In \emph{Proceedings of the 24th ACM international conference on
  Multimedia}, 781--790.

\end{thebibliography}

\end{document}